\documentclass[conference,10pt,letterpaper,twoside]{style/ieeeconf}

\pdfoutput=1

\IEEEoverridecommandlockouts                              
\makeatletter

\let\proof\@undefined
\let\endproof\@undefined

\let\NAT@parse\undefined
\makeatother

\usepackage[numbers, sort, compress]{natbib}
\usepackage{graphicx}
\usepackage{amsmath,amssymb,bm}
\usepackage{subfig}
\usepackage{placeins}

\usepackage{amsthm}  

\usepackage{algorithm}
\usepackage{algorithmic}

\usepackage{multicol}

\usepackage{balance}
\usepackage{mathrsfs}

\usepackage{verbatim}
\usepackage{float}

\usepackage{style/perl_acronyms}
\usepackage{style/perl_math}
\usepackage{style/perl_SIunits}
\usepackage{style/perl_misc}

\usepackage{color}
\usepackage[dvipsnames]{xcolor}

\usepackage{times}
\usepackage{amssymb}
\usepackage{mathtools}
\usepackage[normalem]{ulem}
\usepackage{float}

\usepackage{multicol}
\usepackage[bookmarks=true]{hyperref}

\usepackage{amsthm}
\usepackage{float}
\usepackage{multirow}

\providecommand{\R}{\ensuremath \mathbb{R}}

\providecommand{\C}{\ensuremath \mathbb{C}}
\providecommand{\K}{\ensuremath \mathbb{K}}
\providecommand{\E}{\ensuremath \mathbb{E}}

\providecommand{\Z}{\ensuremath \mathbb{Z}}

\providecommand{\sym}{\text{sym}}
\providecommand{\nnz}{\text{nnz}}

\providecommand{\diag}{\text{diag}}

\providecommand{\mE}{\ensuremath \mathfrak{E}}

\newtheorem{defn}{Definition}
\newtheorem{remark}[defn]{Remark}
\newtheorem{lem}[defn]{Lemma}
\newtheorem{prop}[defn]{Proposition}
\newtheorem{assum}[defn]{Assumption}

\newtheorem{thm}[defn]{Theorem}
\newtheorem{cor}[defn]{Corollary}

\providecommand{\Z}{\mathcal{Z}}

\providecommand{\Z}{\mathcal{Z}}

\providecommand{\I}{\mathcal{I}}
\providecommand{\J}{\mathcal{J}}

\usepackage{marginnote}

\begin{document}
\title{Guaranteed Globally Optimal Planar Pose Graph and Landmark SLAM via Sparse-Bounded Sums-of-Squares Programming}

\author{Joshua~G.~Mangelson,~Jinsun~Liu,~Ryan~M.~Eustice,~and Ram~Vasudevan%
  \thanks{*This work was supported by the \acl{ONR} under awards
    N00014-16-1-2102 and N00014-18-1-2575.}%
  \thanks{J.~Mangelson, J.~Liu, R.~Eustice, and R.~Vasudevan are at the University of Michigan, Ann Arbor, MI 48109, USA. \texttt{\{mangelso, jinsunl, eustice, ramv\}@umich.edu}.}  %
}

\maketitle

\begin{abstract}
  Autonomous navigation requires an accurate model or map of the environment. 
  While dramatic progress in the prior two decades has enabled large-scale \ac{SLAM}, the majority of existing methods rely on non-linear optimization techniques to find the \ac{MLE} of the robot trajectory and surrounding environment. 
   These methods are prone to local minima and are thus sensitive to initialization. 
  Several recent papers have developed optimization algorithms for the Pose-Graph SLAM problem that can certify the optimality of a computed solution.
  Though this does not guarantee {\it a priori} that this approach generates an optimal solution, a recent extension has shown that when the noise lies within a critical threshold that the solution to the optimization algorithm is guaranteed to be optimal. 
To address the limitations of existing approaches, this paper illustrates that the Pose-Graph SLAM and Landmark SLAM can be formulated
  as polynomial optimization programs that are \ac{SOS} convex.
This paper then describes how the Pose-Graph and Landmark SLAM problems can be solved to a global minimum without initialization regardless of noise level using the \ac{Sparse-BSOS} hierarchy. 
This paper also empirically illustrates that convergence happens at the second step in this hierarchy.
In addition, this paper illustrates how this Sparse-BSOS hierarchy can be implemented in the complex domain and empirically shows that convergence happens also at the second step of this complex domain hierarchy.
Finally, the superior performance of the proposed approach when compared to existing SLAM methods is illustrated on graphs with several hundred nodes.
\end{abstract}


\acresetall

\IEEEpeerreviewmaketitle




\section{Introduction}
\label{sec:introduction}

An accurate map of the environment is essential for safe autonomous navigation in the real-world \cite{cadena2016past}. 
An error in the map has the potential to cause loss of life in self-driving car applications or the loss of millions/billions of dollars of assets/time resources when performing underwater or space exploration tasks. 
Despite the importance of accurate mapping, the majority of algorithms used for \ac{SLAM} are prone to local minima and are sensitive to initialization. 
Troublingly, verification of these maps is either performed by visual inspection or not at all.

There has been significant recent interest in developing optimization and estimation algorithms that provide mathematical guarantees on whether a computed solution is or is close to 
the global optimum and is therefore true \ac{MAP} estimate of the map \cite{rosen2016sesync, carlone2016planar, rosen2015convex, carlone2015lagrangian, hu2013towards, liu2012convex}. 
These algorithms either use a relaxation or the dual of the original problem to find a solution. 
As a result these methods either return an approximate solution, are only able to certify the optimally of a solution after it has been computed, or are only able to return the global solution if the graph meets certain requirements related to limits on noise measurement. 
In addition, with the exception of \cite{hu2013towards}, these methods are focused on pose-graph optimization and are unable to handle landmark position measurements or are unable to estimate landmark positions.

\begin{figure}[!tb]%
  \centering%
  \subfloat[Ground Truth - No Noise]{%
    \includegraphics[trim={4.1cm, 8.5cm, 4.5cm, 8.5cm},clip,width=0.5\columnwidth]{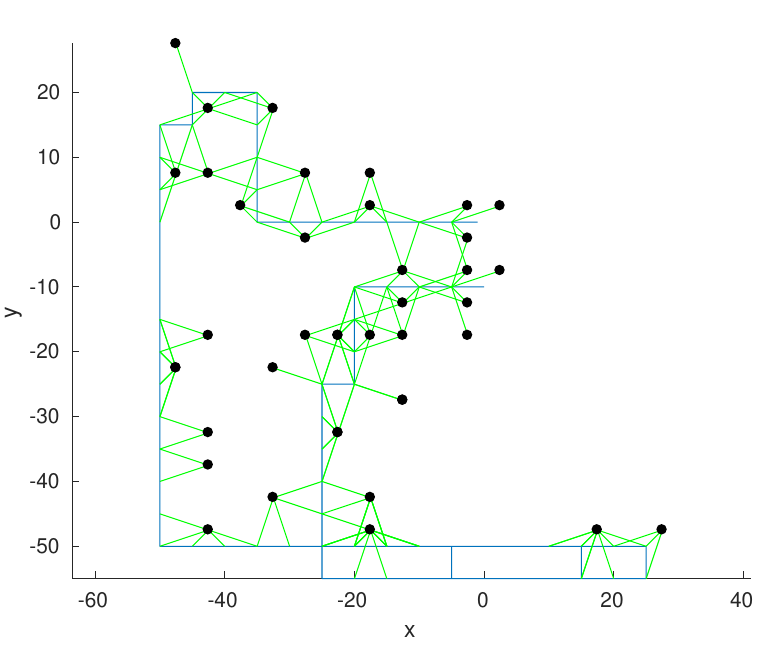}%
    \label{fig:1_gt}%
  }%
  \subfloat[Random Initialization]{%
    \includegraphics[trim={4.1cm, 8.5cm, 4.5cm, 8.5cm},clip,width=0.5\columnwidth]{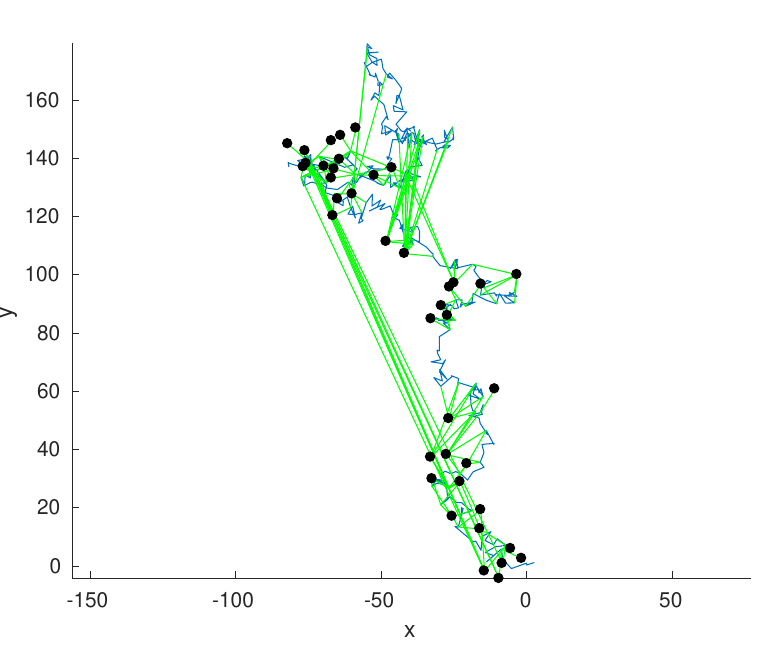}%
    \label{fig:1_random_init}%
  }%
  \hfil%
  \subfloat[Levenberg-Marquardt]{%
    \includegraphics[trim={4.1cm, 8.5cm, 4.5cm, 8.5cm},clip,width=0.5\columnwidth]{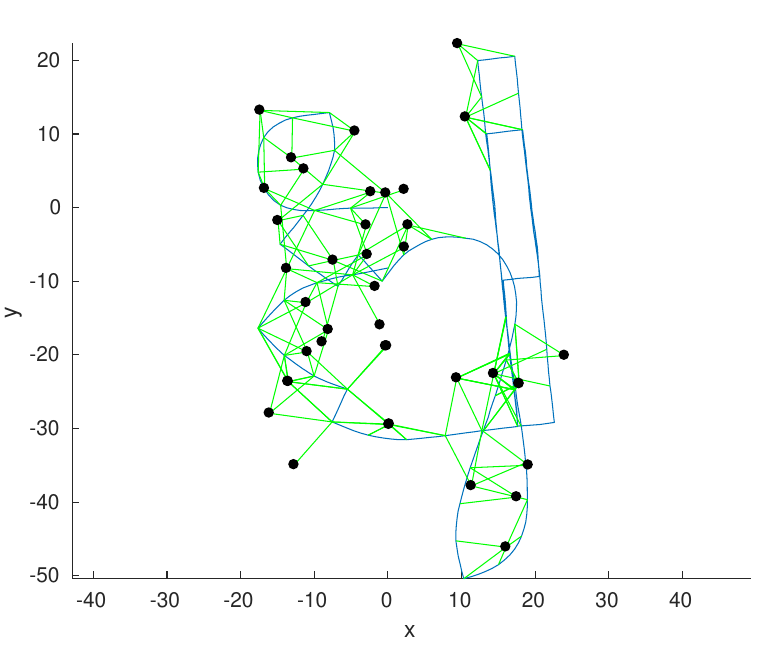}%
    \label{fig:1_LM}%
  }%
  \subfloat[SBSOS-SLAM]{%
    \includegraphics[trim={4.1cm, 8.5cm, 4.5cm, 8.5cm},clip,width=0.5\columnwidth]{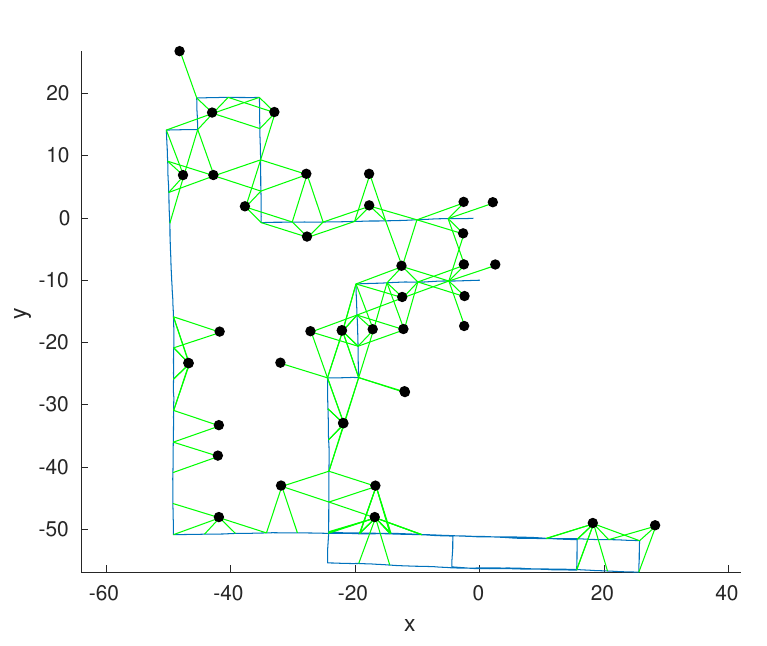}%
    \label{fig:1_SBSOS}%
  }%
  \label{fig:cityTrees430}
  \caption{\small Estimated Landmark SLAM solution for the first 430 nodes(40 landmarks) of the CityTrees10000 dataset \cite{kaess2008isam}. %
    \protect\subref{fig:1_gt} shows the groundtruth pose and landmark positions before noise is added. Levenberg-Marquardt was randomly initialized %
    and becomes trapped in a local minimum. \protect\subref{fig:1_random_init} shows the random initialization and \protect\subref{fig:1_LM} shows the
    LM solution. \protect\subref{fig:1_SBSOS} shows the %
    optimal solution found by SBSOS-SLAM. Our algorithm formulates the Pose Graph and Landmark \ac{SLAM} problems as SOS optimization problems which require no initializations.}
\end{figure}

In the original version of this paper, we argued in error that convergence could happen at the first step of the SBSOS hierarchy. This error arose due to a misapplication of Theorem 3 in \cite{weisser2018sparse}. 
This error was pointed out by several colleagues \cite{brynte2021tightness}. 
To address this mistake, as depicted in Figure \ref{fig:1_SBSOS}, the contributions of this paper are the following:
\begin{enumerate}
\item We formulate the pose graph and landmark planar \ac{SLAM} as polynomial optimization programs.
\item We describe how the \ac{Sparse-BSOS} hierarchy of semidefinite programs (SDP) \acused{SDP} can be used to find its solution \cite{lasserre2017bounded, weisser2018sparse}.
\item We empirically illustrate that convergence of the SLAM problem happens at the second step of the hierarchy
\item We show that we can formulate the Sparse-BSOS hierarchical description of the pose-graph SLAM problem as an equivalent hierarchy of sparse semidefinite programs in the complex domain. 
We empirically illustrate that the pose-graph SLAM problem formulated as a sparse semidefinite program hierarchy over the complex domain converges at the second step of the hierarchy. 
\end{enumerate}

\section{Related Work}
\label{sec:related_work}

\ac{SLAM} refers to the problem of estimating the trajectory of a robotic vehicle over time while simultaneously estimating a model of the surrounding environment \cite{cadena2016past}. 
Initial algorithms used extended Kalman filter and particle filter based methods to simultaneously estimate the position of the robot and the position of observed landmarks in the environment \cite{durrant2006a, bailey2006a, thrun2005probabilistic}, which we refer to as the Landmark SLAM problem. 
Since these methods had challenges scaling to larger datasets, researchers began applying information filter and \ac{MLE} based methods which could exploit sparsity to solve larger instances of the \ac{SLAM} problem. 
To improve the sparsity of the problem, research shifted to solving the Pose Graph SLAM problem wherein the landmarks are marginalized out and only the pose of the robot is optimized over at each time step. 
The majority of modern \ac{SLAM} algorithms seek to find the \ac{MLE} of the robot trajectory through the use of nonlinear estimation based techniques
\cite{lu1997a, eustice2005sparse, kaess2008isam, dellaert2006a}. 
However, the non-linear optimization algorithms used in these methods are dependent on initialization.

Several algorithms leverage theory from the field of convex optimization to overcome this dependence on initialization \cite{boyd2004convex}. 
Optimization over the special euclidean group ($\mathrm{SE}(d)$) has generally been considered a non-convex problem and thus the majority of algorithms rely on some form of convex relaxation to estimate an approximate and sometimes exact solution to the problem. 
For instance the Pose Graph and Landmark \ac{SLAM} problems have been formulated as a non-convex quadratically constrained quadratic program, which was then relaxed into an \ac{SDP} \cite{liu2012convex,hu2013towards}. 
\citet{rosen2015convex} relaxes optimization over the special orthogonal group ($\mathrm{SO}(d)$) to the convex hull of $\mathrm{SO}(d)$ which can be represented using convex
semidefinite constraints. 
Since each of these methods only provide an approximate solution to the \ac{SLAM} problem, they are usually only used as an initial stage and their output is then used to initialize a non-linear optimization method \cite{kaess2008isam, dellaert2006a}.

A number of methods take advantage of Lagrangian Duality to convert the Pose Graph \ac{SLAM} problem into a convex optimization problem that is equivalent to the original optimization problem if the duality gap is zero \cite[Section 5.3.2]{boyd2004convex}.
\citet{carlone2015lagrangian} uses Lagrangian Duality to develop a pair of methods to verify if a computed solution is globally optimal.
\citet{carlone2016planar} applies a similar technique to the planar Pose Graph \ac{SLAM} problem. 
SE-Sync proposed by \citet{rosen2016sesync} extends this prior work and dramatically increases the scalability of the algorithm by taking advantage of a technique called the Riemannian staircase \cite{boumal2015arXiv_reimannian} that enables efficient optimization over semidefinite matrices if the solution has low-rank. 
These methods are only guaranteed to find the globally optimal solution if the measurement noise in the problem lies below a critical threshold and are restricted to the case of Pose Graph \ac{SLAM} where factors are relative pose measurements in $\mathrm{SE}(d)$.
More recently techniques have been proposed to formulate these descriptions of the Pose Graph \ac{SLAM} problem in the complex domain wherein they can be solved more efficiently than similar problems formulated over the real domain \cite{fan2019efficient,fan2020cpl}.



\section{Notation and Preliminaries}

This section defines the notation used throughout the remainder of the paper and presents several preliminary results from the literature which are used in the paper.

Let bold lowercase letters represent vectors and blackboard bold uppercase letters represent sets.
Let $i$ denote the imaginary unit and $\emptyset$ the empty set.
Let $|z| = \sqrt{z_\text{re}^2+z_\text{im}^2}$ denote the magnitude of a complex number $z=z_\text{re}+z_\text{im}\cdot i$ where $z_\text{re}$ and $z_\text{im}$ give the real and imagine parts of $z$.
Let $|\mathbb A|$ denote the cardinality of a set $\mathbb A$, and let $\mathbb A\setminus \mathbb B$ denote the set subtraction between sets $\mathbb A$ and $\mathbb B$.
Let $\{a_j\}_j$ denote the collection of items $a_j$ indexed by $j$.  
Let $\R$, $\C$, $\Z$, and $\Z_+$ denote the sets of real numbers, complex numbers and integers and positive integers, respectively.
Let $\|\bm{v}\|$ denote the Euclidean norm of vector $\bm v$.
Let $\mathbb S_m\subset \R^{m\times m}$ denote the collection of $m$-by-$m$ symmetric real matrices.
Let $\mathbb{S}^1 \subset \C$ denote the circle centered at zero with radius 1 in the complex domain.
Let $\mathbb{T}^n := \underbrace{\mathbb{S}^1 \times \ldots \times \mathbb{S}^1}_n \subset \C^n$ denote the torus. 

Given an arbitrary matrix $A\in\R^{m\times m}$, let $A^\top$ denote its transpose, let $tr(A)$ denote its trace, let $[A]_{(j_1,j_2)}$ denote its $(j_1,j_2)$-th element, let $\nnz(A)$ denote the number of nonzero elements, and let $\sym(A):=\frac{1}{2}(A + A^\top)$. 
Suppose $A$ is a symmetric matrix $A$, then let $A\succeq 0$ denote that the matrix is positive semi-definite and let $A\succ 0$ denote that the matrix is positive definite.
Let $I_m$ and $0_m$ denote the $m$-dimensional identity and zero matrices, respectively.
Let $1_{m_1\times m_2}$ denote the $m_1-$by$-m_2$ matrix with all elements being 1, and $c_m$ denote the $m$-by-$m$ square matrix with all elements being the constant $c$. 
Let $\text{diag}(A)$ denote a row vector that collects diagonal elements of matrix $A$.




\section{Polynomial Optimization SLAM Formulation}
\label{sec:problem_formulation}

This section formulates the Pose Graph and Landmark \ac{SLAM} problems as polynomial optimization programs.

\subsection{Polynomial Optimization}

A polynomial optimization program is an optimization problem of the following form \cite[Section 2.2]{lasserre2017bounded}:
\begin{equation}
  f^* = \underset{\mathbf{x}}{\operatorname{min}} \: \{ f(\mathbf{x}) : \mathbf{x} \in \mathbf{K} \} \label{eq:poly_opt_cost}
\end{equation}
where $f \in \mathbb{R}[\mathbf{x}]$, $\mathbb{R}[\mathbf{x}]$ is the ring of all possible polynomials in the variable $\mathbf{x}=(x_1, \dots, x_N)$, and $\mathbf{K} \subset \mathbb{R}^N$ is the semi-algebraic set
\begin{equation}
  \mathbf{K} = \{ \mathbf{x} \in \mathbb{R}^N : 0 \leq g_j(\mathbf{x}) \leq 1, j=1, \dots, M\}, \label{eq:basic_semi_algebraic_set}
\end{equation}
for polynomials $g_j \in \mathbb{R}[\mathbf{x}], j=1, \dots, M$.

\subsection{Pose Graph SLAM}
\label{sec:pose_graph_SLAM}

In planar Pose Graph \ac{SLAM}, one estimates the pose of the robot, $(\mathbf{R}_i, \mathbf{t}_i)\in\mathrm{SE}(2)$ with respect to a static global reference frame at each time steps $i \in \{1, \dots, n\}$, by minimizing the error in a set of $m_{Rel}$ relative pose measurements $(\bar{\mathbf{R}}_{ij}, \bar{\mathbf{t}}_{ij})\in\mathrm{SE}(2)$. 
The set of available measurements can be represented by the set of edges, $E = \{i_k, j_k\}_{k=1}^{m_{Rel}} \subset \{1,\dots,n\} \times \{1,\dots,n\}$, in the corresponding factor graph. 
We denote the pose of the robot at time step $i$ by the matrix $\mathbf{H}_{i} =[\mathbf{R}_i | \mathbf{t}_i]$ and the relative pose measurement that relates the pose of the robot at time steps $i$ and $j$ by $\bar{\mathbf{H}}_{ij}=[\bar{\mathbf{R}}_{ij}|\bar{\mathbf{t}}_{ij}]$.
We assume that each $\bar{\mathbf{R}}_{ij}$ and $\bar{\mathbf{t}}_{ij}$ are conditionally independent given the true state, that $\bar{\mathbf{R}}_{ij}\sim \text{Langevin}(\mathbf{R}_{ij}, \omega^2_{\mathbf{R}_{ij}})$, and that $\bar{\mathbf{t}}_{ij} \sim \mathcal{N}(\mathbf{t}_{ij}, \Omega_{\mathbf{t}_{ij}}^{-1})$, where $(\mathbf{R}_{ij}, \mathbf{t}_{ij})$ is the true relative pose, $\omega^2_{\mathbf{R}_{ij}}$ is the concentration parameter of the Langevin Distribution, and $\Omega_{\mathbf{t}_{ij}} = \text{blkdiag}(\omega^2_{x_{ij}}, \omega^2_{y_{ij}})$ is the information matrix of $\bar{\mathbf{t}}_{ij}$.
Note $\text{blkdiag}(\omega^2_{x_{ij}}, \omega^2_{y_{ij}})$ denotes a block diagonal matrix whose diagonal elements are equal to $\omega^2_{x_{ij}}$ and $\omega^2_{y_{ij}}$.

Under these assumptions, the \ac{MLE} solution to the planar pose graph SLAM problem is equivalent to:
\begin{align}
  \underset{\mathbf{H}_1, \dots, \mathbf{H}_n \in \mathrm{SE}(2)}{\operatorname{\text{argmin}}} \: \sum\limits_{(i,j)\in E} &\omega^2_{\mathbf{R}_{ij}} \left| \left| \mathbf{R}_j - \mathbf{R}_i \bar{\mathbf{R}}_{ij}\right| \right|_F^2 +  \label{eq:se2_pose_factor}\\
  &~~\left| \left| \mathbf{t}_j - \mathbf{t}_i - \mathbf{R}_i \bar{\mathbf{t}}_{ij}\right| \right|_{\Omega_{\mathbf{t}_{ij}}}^2 \nonumber
\end{align}
where $||\cdot||_F$ is the Frobenius norm and $||\mathbf{x}||^2_{\Omega} = \mathbf{x}^\top \Omega \mathbf{x}$ for $x \in \mathbb{R}^2$ \cite{rosen2016sesync, briales2017cartan}.

Note that $\mathbf{R}_{i} \in \mathrm{SO}(2)$ for each $i\in \{1,\dots,n\}$. 
$\mathrm{SO}(2)$ can be defined as follows:
\begin{align}
  \mathrm{SO}(2) =
  \left \{ \mathbf{R} =
  \left [
  \begin{array}{cc}
    c & -s \\
    s & c
  \end{array}
        \right ] \in \mathbb{R}^{2\times 2}  ~\bigl\vert~
        c^2 + s^2 = 1
        \right \}.
        \label{eq:def_so2}
\end{align}
This definition allows us to parameterize $\mathbf{H}_i$ and $\mathbf{H}_{ij}$ using $(c_i, s_i, x_i, y_i)$ and
$(c_{ij}, s_{ij}, x_{ij}, y_{ij})$ respectively,
\begin{align}
\mathbf{H}_i = \left[ \begin{array}{ccc}
                        c_i & -s_i & x_i \\
                        s_i & c_i  & y_i \\
                      \end{array} \right],
\mathbf{H}_{ij} = \left[ \begin{array}{ccc}
                        c_{ij} & -s_{ij} & x_{ij} \\
                        s_{ij} & c_{ij}  & y_{ij} \\
                      \end{array} \right],  
\end{align}
as long as we enforce that $c_i^2 + s_i^2 = 1$.
To simplify this notation, we define the sets $\mathbf{c}=\{c_1,\dots,c_n\}$, $\mathbf{s}=\{s_1,\dots,s_n\}$, $\mathbf{x}=\{x_1,\dots,x_n\}$, and $\mathbf{y}=\{y_1,\dots,y_n\}$.

If we evaluate the norms in \eqref{eq:se2_pose_factor} under this parameterization, then  \eqref{eq:se2_pose_factor} is equivalent to 
\begin{align}
  &\underset{\mathbf{c}, \mathbf{s}, \mathbf{x}, \mathbf{y}}{\operatorname{\text{argmin}}} \: \sum\limits_{(i,j)\in E} f^\mathbf{H}_{ij}(c_i, s_i, x_i, y_i, c_j, s_j, x_j, y_j)
                                                                                                                              \label{eq:pose_slam_mle_factored} \\
  &~~~~~\text{s.t.} ~~~ c_i^2 + s_i^2 = 1, \quad \forall i\in\{1,\dots,n\}, \nonumber
\end{align}
with,
\begin{align}
  f^{\mathbf{H}}_{ij}(c_i, &s_i, x_i, y_i, c_j, s_j, x_j, y_j) = f^{Rot}_{ij}(c_i, s_i, c_j, s_j) + \\
  &+ f^{Tran}_{ij}(c_i, s_i, x_i, y_i, x_j, y_j), \nonumber
\end{align}
where
\begin{align}
  &f^{Rot}_{ij}(c_i, s_i, c_j, s_j) =  \omega_{\mathbf{R}_{ij}}^2 (c_j - c_i c_{ij} + s_i s_{ij})^2 + \label{eq:rot_polynomial_cost} \\ 
  & + \omega_{\mathbf{R}_{ij}}^2 (-s_j + c_i s_{ij} + s_i c_{ij})^2 + \omega_{\mathbf{R}_{ij}}^2 (s_j - s_i c_{ij} + \\
  &  - c_i s_{ij})^2 + \omega_{\mathbf{R}_{ij}}^2 (c_j + s_i s_{ij} - c_i c_{ij})^2, \nonumber
\end{align}%
and
\begin{align}
  f^{Tran}_{ij}(c_i, &s_i, x_i, y_i, x_j, y_j) = \omega_{x_{ij}}^2 (x_j - c_i x_{ij} + \label{eq:tran_polynomial_cost} \\
  & + s_i y_{ij} - x_i)^2 + \omega_{y_{ij}}^2 (y_j - s_i x_{ij} - c_i y_{ij} - y_i)^2. \nonumber
\end{align}
Note that the cost is a polynomial in the space $\mathbb{R}[\mathbf{c}, \mathbf{s}, \mathbf{x}, \mathbf{y}]$ and that each individual term $f^\mathbf{H}_{ij} \in \mathbb{R}[c_i, s_i, x_i, y_i, c_j, s_j, x_j, y_j]$.
We can also rewrite $c_i^2 + s_i^2 = 1$ as $0 \leq 1 - c_i^2 - s_i^2 \leq 1$ and $ 0 \leq 2 - c_i^2 - s_i^2 \leq 1$ for each $i \in \{1,\ldots,n\}$.
This parameterization allows us to rewrite \eqref{eq:pose_slam_mle_factored} as a polynomial optimization problem in the form described in \eqref{eq:poly_opt_cost} and \eqref{eq:basic_semi_algebraic_set}, where $M=2n$:
\begin{align}
  \underset{\mathbf{c}, \mathbf{s}, \mathbf{x}, \mathbf{y}}{\operatorname{\text{argmin}}} \: &\sum\limits_{(i,j)\in E} f^\mathbf{H}_{ij}(c_i, s_i, x_i, y_i, c_j, s_j, x_j, y_j)
                                                                                                                              \label{eq:pose_slam_poly_opt} \\
  ~~~~~\text{s.t.} ~~~ &0 \leq 1 - c_i^2 - s_i^2 \leq 1, ~ \forall i\in\{1,\dots,n\}, \nonumber \\
  &0 \leq 2 - c_i^2 - s_i^2 \leq 1, ~ \forall i\in\{1,\dots,n\}. \nonumber    
\end{align}

\subsection{Landmark SLAM}
\label{sec:landmark_SLAM}

In Landmark \ac{SLAM}, one estimates both the pose of the robotic vehicle at each time step, $(\mathbf{R}_i, \mathbf{t}_i)\in\mathrm{SE}(2)$, as well as the position of observed landmarks, $\mathbf{l}_\ell=[l^x_{\ell}, l^y_\ell]^\top\in\mathbb{R}^2$ for each $\ell \in \{1,\dots,w\}$, given both relative pose measurements $(\bar{\mathbf{R}}_{ij}, \bar{\mathbf{t}}_{ij})\in\mathrm{SE}(2)$ and landmark position observations $\bar{\mathbf{l}}_{i\ell} = [x_{i\ell}, y_{i\ell}]^\top \in \mathbb{R}^2$ that measure the position of landmark with respect to the local coordinate frame of the robot at the time step that it was observed. 
Let $L=\{i_k,\ell_k\}_{k=1}^{m_\ell} \subset \{1,\dots,n\}\times\{1,\dots,w\}$ identify the set of landmark position measurements where $m_\ell$ is the number of landmark measurements and let $\mathbf{l_x}=\{l_1^x,\dots,l_w^x\}$ and $\mathbf{l_y}=\{l_1^y,\dots,l_w^y\}$.
We assume that the relative pose measurements are distributed
according to the structure defined in the previous section and that
$\bar{\mathbf{l}}_{i\ell} \sim \mathcal{N}(\mathbf{l}^i_{i\ell}, \Omega_{\mathbf{l}_{i\ell}}^{-1})$ where $\mathbf{l}^i_{i\ell}$ is the true position of the landmark $\ell$ with respect to the true pose $\mathbf{H}_i$ and $\Omega_{\mathbf{l}_{i\ell}}=\text{blkdiag}(\omega_{x_{i\ell}}^2, \omega_{y_{i\ell}}^2)$ is the information matrix of $\bar{\mathbf{l}}_{i\ell}$.

Under these assumptions, the \ac{MLE} solution to the Landmark \ac{SLAM} problem can be written as follows:
\begin{align}
  \underset{\mathbf{c}, \mathbf{s}, \mathbf{x}, \mathbf{y}, \mathbf{l_x}, \mathbf{l_y}}{\operatorname{\text{argmin}}} \: &\sum\limits_{(i,j)\in E} f^\mathbf{H}_{ij}(c_i, s_i, x_i, y_i, c_j, s_j, x_j, y_j) ~+
                                                                                               \label{eq:landmark_slam_poly_opt} \\
  &~~~~ + \sum\limits_{(i,\ell)\in L} f^{Land}_{i\ell}(c_i, s_i, x_i, y_i, l_\ell^x, l_\ell^y) \nonumber \\
  ~~~~~\text{s.t.} ~~~ &0 \leq 1 - c_i^2 - s_i^2 \leq 1, ~ \forall i\in \{1,\dots,n\}, \nonumber \\
  &0 \leq 2 - c_i^2 - s_i^2 \leq 1, ~ \forall i\in \{1,\dots,n\}. \nonumber    
\end{align}
with, 
\begin{align}
  f^{Land}_{i\ell}(c_i, &s_i, x_i, y_i, l_\ell^x, l_\ell^y) = \omega_{x_{i\ell}}^2 (l_\ell^x - c_i x_{i\ell} + \label{eq:landmark_tran_polynomial_cost} \\
  & + s_i y_{i\ell} - x_i)^2  + \omega_{y_{i\ell}}^2 (l_\ell^y - s_i x_{i\ell} - c_i y_{i\ell} - y_i)^2. \nonumber
\end{align}
Note that the cost of the optimization problem in \eqref{eq:landmark_slam_poly_opt} is a polynomial in the space $\mathbb{R}[\mathbf{c}, \mathbf{s}, \mathbf{x}, \mathbf{y}, \mathbf{l_x}, \mathbf{l_y}]$ while $f^{Land}_{i\ell}\in \mathbb{R}[c_i, s_i, x_i, y_i, l_x, l_y]$. Also note that the constraints are the same as in \eqref{eq:pose_slam_poly_opt} and thus, \eqref{eq:landmark_slam_poly_opt} is a polynomial optimization problem of the form defined in \eqref{eq:poly_opt_cost} and \eqref{eq:basic_semi_algebraic_set}.


\section{Sparse Bounded Sum-of-Squares Programming}
\label{sec:SBSOS}

Polynomial optimization problems in general are non-convex, however, they can be approximated and sometimes solved exactly by solving a hierarchy of convex relaxations of the problem \cite{lasserre2009moments}. 
A variety of such convex relaxations hierarchies exist. 
This section covers a pair of such hierarchies. 
The first is called the \ac{BSOS} hierarchy and consists of a sequence of \ac{SDP} relaxations that can be used to find the globally optimal solution to small polynomial optimization problems that meet certain conditions \cite{lasserre2017bounded}. 
The second is called \ac{Sparse-BSOS} and enables us to leverage the sparsity inherent in SLAM problems to solve larger problem sizes than is possible using \ac{BSOS} \cite{weisser2018sparse}. 
We conclude the section by describing the conditions that the cost and constraints that a polynomial optimization must satisfy for the first step of either hierarchy to converge exactly to the global optimum.

\subsection{Bounded Sum-of-Squares}
\label{sec:BSOS}

\ac{SOS} programming is concerned with finding solutions to polynomial optimization problems as in \eqref{eq:poly_opt_cost}. 
If $\mathbf{x}$ did not have to lie within the semi-algebraic set $\mathbf{K}$, solving the following problem would be equivalent to solving \eqref{eq:poly_opt_cost}:
\begin{equation}
  t^* = \underset{t \in \mathbb{R} }{\operatorname{sup}} \: \{t~ |~ f(\mathbf{x}) - t \geq 0, \forall~\mathbf{x}\}.
\end{equation}
If instead one had constraints $g_j$ that bound the feasible space of the variable $\mathbf{x}$ to $\mathbf{K}$, then one would need to enforce that $f(\mathbf{x}) - t \geq 0,~\forall~\mathbf{x} \in \mathbf{K}$.
At the same time, one would have to enforce it in a way that enabled $f - t$ to get as close to zero as possible at the optimal solution. 
Suppose we could optimize over a function $h$ and also strictly enforce that it be non-negative on $\mathbf{K}$. 
Then, by enforcing that $f(\mathbf{x}) - t - h(\mathbf{x}) \geq 0$, for all $x$, we would equivalently enforce that $f(\mathbf{x}) - t \geq h(\mathbf{x}) \geq 0$ on $\mathbf{K}$ and we would be able to optimize over $h$ to minimize the gap between $f$ and $t$ on $\mathbf{K}$.

To apply this approach using numerical optimization, one would first need to know whether it was computationally tractable to enforce positivity of $h$ on $K$. 
Assuming that $0 \leq g_j(x) \leq 1$ for all $x \in \mathbf{K}$ and $\mathbf{K}$ is compact, one can prove that if a polynomial $h$ is strictly positive on $\mathbf{K}$, then $h$ can be represented as 
\begin{align}
  h(\mathbf{x}, \boldsymbol{\lambda}) = \sum\limits_{\alpha, \beta \in \mathbb{N}^M} \lambda_{\alpha\beta} \prod_j\left(g_j(\mathbf{x})^{\alpha_j}(1 - g_j(\mathbf{x}))^{\beta_j}\right),
  \label{eq:bsos_thm_1}
\end{align}
for some (finitely many) nonnegative scalars $\boldsymbol{\lambda} = (\lambda_{\alpha\beta}$) \cite[Theorem 1]{lasserre2017bounded}\footnote{Note that the theorem as presented requires the set $\{1,g_1,\ldots,g_M\}$ to generate $\mathbf{K}$, but since $\mathbf{K}$ is compact, one can always add a redundant linear constraint to the set to satisfy this requirement.}. 
Conversely, any polynomial that can be written in the form defined in \eqref{eq:bsos_thm_1} is also positive on $\mathbf{K}$. 
This leads to a hierarchy of relaxations in which each relaxation bounds the number of monomial terms used to represent $h$ \cite[Theorem 2]{lasserre2017bounded}.

Let $N^{2M}_d = \{(\alpha, \beta) | \alpha, \beta \in \mathbb{N}^M, |\alpha| + |\beta| \leq d\}$ where the absolute value denotes the sum and
\begin{align}
h_{d}(\mathbf{x}, \boldsymbol{\lambda}) := \sum\limits_{(\alpha, \beta) \in N_d^{2M}} \hspace*{-0.25cm} \lambda_{\alpha\beta} \prod_{j=1}^M g_j(\mathbf{x})^{\alpha_j}(1 - g_j(\mathbf{x}))^{\beta_j},
\end{align}
where $\boldsymbol{\lambda} = (\lambda_{\alpha\beta}), (\alpha, \beta) \in \mathbb{N}^{2M}_d$. 
By choosing $d$, one can bound the number of monomial terms that are used to represent $h_d$ and by optimizing over $\lambda$, one can optimize
over the specific polynomial. 
By constraining $\boldsymbol{\lambda}$ to be non-negative, one can enforce that $h_d$ be strictly positive on $\mathbf{K}$.

Now one can solve the following optimization problem:
\begin{align}
  t^* = \underset{t, \boldsymbol{\lambda} }{\operatorname{sup}} \{ t | f(\mathbf{x}) - t - h_d(\mathbf{x}) \geq 0, \forall~\mathbf{x}, \boldsymbol{\lambda} \geq 0\}.
  \label{eq:bsos_infinite_dimensional}
\end{align}
However, optimizing over the space of all positive polynomials is computationally intractable.
Instead, one can relax the problem again and optimize over the space of \ac{SOS} polynomials up to a fixed degree since
\ac{SOS} polynomials are guaranteed to be positive and can be represented using a positive semidefinite matrix \cite[Chapter 2]{lasserre2009moments}.
Let $\Sigma[\mathbf{x}] \subset \mathbb{R}[\mathbf{x}]$ represent the space of \ac{SOS} polynomials and let $\Sigma[\mathbf{x}]_k \subset \mathbb{R}[\mathbf{x}]_{2k}$ represent the space of \ac{SOS} polynomials of degree at most $2k$.
By fixing $k \in \mathbb{N}$, one arrives at the following \ac{BSOS} family of convex relaxations:
indexed by $d\in \mathbb{N}$:
\begin{align}
   q^k_d= \underset{t, \boldsymbol{\lambda} }{\operatorname{sup}} \{ t | f(\mathbf{x}) - t - h_d(\mathbf{x}) \in \Sigma[\mathbf{x}]_k, \boldsymbol{\lambda} \geq 0\}.
  \label{eq:bsos_hierarchy}
\end{align}
Each of these optimization programs can be implemented as an \ac{SDP} and provides a lower bound on the solution
to \eqref{eq:bsos_infinite_dimensional}. 
Additionally, it can be shown that under certain assumptions as $d \to \infty$, $q^k_d \to f^*$ \cite[Theorem 2]{lasserre2017bounded}. 
While this is useful for small problems, as the number of variables increases or for larger values of $d$ and $k$, the runtime and memory usage of the optimization makes the use of this method infeasible \cite[Section 3]{lasserre2017bounded}. 
To address this challenge, we take advantage of sparsity in the optimization problem to dramatically scale problem size.

\subsection{Sparse Bounded Sum-of-Squares}
\label{sec:Sparse-BSOS}

The \ac{Sparse-BSOS} hierarchy takes advantage of the fact that for many optimization problems, the variables and constraints exhibit structured sparsity. 
It does this by splitting the variables in the problem into $p$ blocks of variables and the cost into $p$ associated terms, such that the number of variables and constraints relevant to each block is small \cite{weisser2018sparse}.

Given $I \subset \{1, \dots, N\}$, let $\mathbb{R}[\mathbf{x}; I]$ denote the ring of polynomials in the variables $\{x_i : i \in I\}$. 
Specifically \ac{Sparse-BSOS} assumes that the cost and constraints satisfy the following assumption:
\begin{assum}[Running Intersection Property (RIP)]\label{assum: rip-sos}
There exists $p\in\mathbb{N}$ and $I_\ell \subseteq \{1, \dots, N\}$ and $J_\ell \subseteq \{1, \dots, M\}$ for all $\ell \in \{1, \dots, p\}$ such that:
\begin{itemize}
\item $f = \sum_{\ell=1}^p f^\ell$, for some $f^1, \dots, f^p$, such that \\$~~~f^\ell \in \mathbb{R}[\mathbf{x}, I_\ell] $ for each $ \ell \in \{1, \dots, p\}$,
\item $g_j \in \mathbb{R}[\mathbf{x}, I_\ell]$ for each $j \in J_\ell$ and $\ell \in \{1, \dots, p\}$,
\item $\cup_{\ell=1}^p I_\ell = \{1, \dots, N\}$,
\item $\cup_{\ell=1}^p J_\ell = \{1, \dots, M\}$,
\item for all $\ell = 1, \dots, p - 1$, there is an $s \leq \ell$ such that $(I_{\ell+1} \cap \cup_{r=1}^\ell I_r) \subseteq I_s$.
\end{itemize}
\end{assum}
\noindent In particular, $I_\ell$ denotes the variables that are relevant to $\ell$-th block and $J_\ell$ denotes the associated relevant constraints. 
Intuitively, these blocks allow one to enforce positivity over a smaller set of variables which can reduce the computational burden while trying to solve this optimization problem. 

We can use these definitions to define the \ac{Sparse-BSOS} hierarchy that builds on the hierarchy defined in the previous section.
Let $N^\ell := \{(\alpha, \beta) : \alpha, \beta \in \mathbb{N}_0, \text{supp}(\alpha)\cup\text{supp}(\beta) \subseteq J_\ell\}$, where $\mathbb{N}_0$ is the set of natural numbers including $0$ and $\text{supp}(\alpha) := \{j \in \{1, \dots, M\} : \alpha_j \neq 0 \}$.
Now let $N^\ell_d := \{(\alpha, \beta) \in N^\ell : \sum_j(\alpha_j + \beta_j) \leq d\}$, with $d \in \mathbb{N}$
and let 
\begin{align}
h^{\ell}_{d}(\mathbf{x}, \boldsymbol{\lambda}^\ell) := \sum\limits_{(\alpha, \beta) \in N_d^{\ell}} \lambda^\ell_{\alpha\beta} \prod_{j=1}^M g_j(\mathbf{x})^{\alpha_j}(1 - g_j(\mathbf{x}))^{\beta_j},
\end{align}
where $\boldsymbol{\lambda}^\ell \in \mathbb{R}^{|N^\ell_d|}$ is the vector of scalar coefficients $\lambda_{\alpha\beta}^\ell$. $h^{\ell}_{d}$ is again positive on $\mathbf{K}$ as long as the elements of $\boldsymbol{\lambda}$ are positive. 
If we again fix $k \in \mathbb{N}$, we can define a family of optimization problems indexed by $d\in\mathbb{N}$ as shown in \eqref{eq:sbsos_hierarchy}, where $d_{\max}$ is defined on page 7 of \cite{weisser2018sparse}. 
\begin{figure*}
\begin{align}
  &q^k_d= \underset{\substack{t, \boldsymbol{\lambda}^1, \dots, \boldsymbol{\lambda}^p, \\ f^1, \dots, f^p}}{\operatorname{sup}} \left\{ t | f^{\ell}(\mathbf{x}) - h^{\ell}_d(\mathbf{x}, \boldsymbol{\lambda}^{\ell}) \in \Sigma[\mathbf{x}; I_{\ell}]_k, \ell = 1, \dots, p, \label{eq:sbsos_hierarchy} \right.\\
    &\left.~~~~~~~~~~~~~~~~~~~~~~f(\mathbf{x}) - t = \sum\limits_{\ell=1}^p f^{\ell}(\mathbf{x}), \boldsymbol{\lambda}^{\ell} \in \mathbb{R}^{|N_d^{\ell}|} \geq 0, t \in \mathbb{R}, f^{\ell} \in \mathbb{R}[\mathbf{x}; I_{\ell}]_{d_{max}}  \right\} . \nonumber
\end{align}
\end{figure*}

This hierarchy of relaxations is called the \ac{Sparse-BSOS} hierarchy and each level of the hierarchy can be implemented as an \ac{SDP}.
In addition, if RIP is satisfied, $0 \leq g_j(x) \leq 1$ for all $x \in \mathbf{K}$, and $\mathbf{K}$ is compact, then as $d \to \infty$, the sequence of optimization problems defined in \eqref{eq:sbsos_hierarchy} also converges to $f^*$\cite[Theorem 2]{weisser2018sparse}.
In addition a rank condition can be used to detect finite convergence \cite[Lemma 4]{weisser2018sparse}. 
Importantly in particular cases, one can show that this optimization problem can be solved exactly when $d = 1$.





\section{SBSOS-SLAM}
\label{sec:SBSOS-SLAM}

We take advantage of the \ac{Sparse-BSOS} relaxation hierarchy to solve the Pose Graph and Landmark \ac{SLAM} problems.
In this section we talk about how we can enforce the RIP. 
We conclude this section with a discussion of implementation.

\subsection{Satisfying the Running Intersection Property}

The nature of the \ac{SLAM} problem exhibits a large amount of sparsity \cite{ eustice2005sparse, kaess2008isam, dellaert2006a}. 
However, to take advantage of the guarantees incumbent to the \ac{Sparse-BSOS} hierarchy, we need to satisfy the RIP. 
          An odometry chain forming the backbone of a \ac{SLAM} graph inherently satisfies this property, however incorporating loop closures can make satisfying this assumption challenging. A better grouping increases the sparsity of the optimization problem and leads to faster solutions, but finding the optimal selection of blocks $I_\ell$ is NP-hard. 
We used the heuristic algorithm defined in \cite{smail2017junction} to generate a sequence of variable groupings for the current implementation.

\subsection{Implementation and Computational Scaling}

For the experiments and development presented in this paper, we modified the code base released with \cite{weisser2018sparse} to formulate the problem as defined earlier on in the paper. 
In addition, we modified the code to convert the problem to a format where we could use the \ac{SDP} solver within the optimization library Mosek \cite{mosek_c_api}. 

\ac{SOS} programming optimizes over polynomials which can become ill-conditioned when optimization occurs over a large domain \cite[Section 4]{weisser2018sparse}. 
Data can be scaled to address this problem, however if the optimization problem still remains poorly scaled then the optimization solver will warn the user that the problem cannot be satisfactorily solved.


\section{Formulating the Rotational Averaging Problem as a sparse SDP in the Complex Domain}

This section begins by describing how to formulate the Pose Graph problem without translation in the complex domain as an SDP and illustrates that it satisfies the RIP sparsity structure. 
We then illustrate how this complex domain SDP can be formulated as a smaller, sparse complex domain SDP that can be more readily solved.


\subsection{Rotational Averaging Problem and RIP}
In this manuscript we aim for solving the $n$-pose \emph{Rotational Averaging Problem}
\begin{align*}
    \inf_{\bm c, \bm s\in\R^n}  \quad & f(\bm{c,s})   \hspace{3cm}\texttt{(RAP)}\\
    \text{s.t.} \quad~~ & c_j^2+s_j^2 = 1, ~~\forall j=1,2,\cdots,n
\end{align*}
where $\bm{c} = [c_1,\cdots,c_n]^\top$ and $\bm{s} = [s_1,\cdots,s_n]^\top$.
For simplicity define $\mathcal C:=\{(\bm{c,s})\mid c_j^2+s_j^2 = 1, ~\forall j=1,2,\cdots,n\}\subset\R^{2n}$ the feasible set of \texttt{(RAP)}.
The \emph{rotational averaging cost} reads
\begin{align}
    f(\bm{c,s}) = &\sum_{(j_1,j_2)\in \E} \big( 2\Omega_{j_1j_2}^2(c_{j_2}-c_{j_1}c_{j_1j_2}+s_{j_1}s_{j_1j_2})^2 +\nonumber \\
    &\hspace{1.1cm} + 2\Omega_{j_1j_2}^2(s_{j_2}-s_{j_1}c_{j_1j_2}-c_{j_1}s_{j_1j_2})^2 \big)\\
    =&\sum_{(j_1,j_2)\in \E} f^\text{rot}_{j_1j_2}(c_{j_1},s_{j_1},c_{j_2},c_{j_2})
\end{align}
where $\E\subset\{1,\cdots,n\}\times\{1,\cdots,n\}$ is the collection of edges in the pose graph, $\{\Omega_{j_1j_2}\}_{(j_1,j_2)\in \E}$ is the set of concentration parameters of the Langevin Distribution, and constants $c_{j_1j_2}$ and $s_{j_1j_2}$ satisfy $c_{j_1j_2}^2 + s_{j_1j_2}^2=1$ for all $(j_1,j_2)\in\E$.

Let $\R[(\bm{c,s})]$ denote the ring of real polynomials over $\bm c$ and $\bm s$.
Given $\I\subseteq\{1,2,\cdots,n\}$, let $\R[(\bm {c, s}); \I]$ denote the ring of polynomials in variables $\{c_{\texttt i}: \texttt i\in\I\}$ and $\{s_{\texttt i}: \texttt i\in\I\}$.
Using the edges of the pose graph, one can show that the rotational averaging cost satisfies a specific sparsity structure similar as in \cite[Assumption 1]{weisser2018sparse}: 
\begin{assum}
\label{assum: rip}
There exists $p\in\Z_+$ and $\I_\ell\subseteq\{1,2,\cdots,n\}$ for all $\ell\in\{1,2,\cdots,p\}$ such that:
\begin{itemize}
    \item There exists $f^1(\cdot),\ldots,f^p(\cdot)$ with $f^\ell(\cdot)\in\R[(\bm{c,s});\I_\ell]$ for each $\ell\in\{1,2,\cdots,p\}$ such that $f(\cdot) = \sum_{\ell=1}^p f^\ell(\cdot)$,
    \item $\cup_{\ell=1}^p\I_\ell = \{1,2,\cdots,n\}$,
    \item for each $\ell \in \{1,2,\cdots,p-1\}$, there exists an $\ell_0\leq\ell$ such that $(\I_{\ell+1}\cap\cup_{\ell_1=1}^\ell\I_{\ell_1})\subseteq \I_{\ell_0}$ \emph{(Running Intersection Property)}.
\end{itemize}
\end{assum}
Note, we do not need to assume index set $\J_\ell$ over constraints as in \cite[Assumption 1]{weisser2018sparse} because we can simply set $\J_\ell = \I_\ell$ for all $\ell$ due to the fact that every constraint in \texttt{(RAP)} depends on only one pose.

\subsection{Formulating the Rotational Averaging Problem in the Complex Domain}

For each $j\in\{1,\cdots,n\}$, let the pair $(c_j,s_j)$  represent the pair of $\sin(\omega_j)$ and $\cos(\omega_j)$ for some angle $\omega_j$, i.e.
\begin{equation}
    c_j = \frac{e^{i\omega_j} + e^{-i\omega_j}}{2} \hspace*{0.5cm} ~ s_j = \frac{e^{i\omega_j} - e^{-i\omega_j}}{2i}.
\end{equation}
Equivalently one can let
\begin{equation}
\label{eq: cs2z}
    c_j = \frac{z_j + z_j^{-1}}{2} \hspace*{0.5cm} ~ s_j = \frac{z_j - z_j^{-1}}{2i}
\end{equation}
for some complex number $z_j\in\C$ such that $|z_j| = 1$.
Let $\nu\in\R$ denote an arbitrary value that is strictly smaller than the minimum value $\nu^*\in\R$ of $f(\bm{c,s})$ over $\mathcal C$. 
Then $f(\bm{c,s})-\nu>0$ over  $\mathcal C$, and we can transfer $f(\bm{c,s})-\nu$ into the complex domain via \eqref{eq: cs2z} to create a Laurent polynomial \cite[Section 1.1]{schmidt1960toeplitz} $R$ as:
\begin{equation}
     R(\bm z) = \sum_{\bm k\in\K} r_{\bm k} \bm{z^k}
\end{equation}
with $|z_j|=1$ for all $j\in \{ 1,\ldots, n\}$ where $\bm{z} = [z_1,\cdots, z_n]^\top\in\C^n$ and $\bm k=[k_1\cdots,k_n]^\top\in\K\subset \Z^n$. 
We postpone the specification of how $\K$ is decided to the next subsection, and the following lemma holds.

\begin{lem}
\label{lem: r0pos}
If $R(\bm z)>0$ for all $\bm{z} \in \mathbb{T}^n$, then the constant term $r_{\bm{0}}$ of $R$ is real and strictly positive. 
\end{lem}

\begin{proof}
To prove this result, we first describe how to transform $f(\bm{c,s})$ into $R(\bm{z})$. 
Then we identify a correspondence between  $r_{\bm{0}}$ and the Inverse Z-Transform of $R$. 
We use this correspondence to prove the desired result. 

Expanding $f^\text{rot}_{j_1j_2}(\cdot)$ we get
\begin{align}
\label{eq: f_rot_j1j2}
    f^\text{rot}_{j_1j_2}(c_{j_1},s_{j_1},c_{j_2},c_{j_2}) =  & 2\Omega_{j_1j_2}^2 \big(c_{j_1}^2 - 2c_{j_1j_2}c_{j_1}c_{j_2} +c_{j_2}^2+\nonumber\\
    &- 2s_{j_1j_2}c_{j_1}s_{j_2} + 2s_{j_1j_2}c_{j_2}s_{j_1} + \nonumber\\
    &+s_{j_1}^2 - 2c_{j_1j_2}s_{j_1}s_{j_2} + s_{j_2}^2\big).
\end{align}
Applying \eqref{eq: cs2z}, $f^\text{rot}_{j_1j_2}$ transforms into
\begin{align}
\label{eq:R_rot_j1j2}
    R^\text{rot}_{j_1j_2}(z_{j_1},z_{j_2}) = &  2\Omega_{j_1j_2}^2 \big(2 + (-c_{j_1j_2}+s_{j_1j_2}\cdot i)z_{j_1}^{-1}z_{j_2} +\nonumber\\
    &+(-c_{j_1j_2}-s_{j_1j_2}\cdot i)z_{j_1}z_{j_2}^{-1}\big), 
\end{align}
in which a constant $4\Omega_{j_1j_2}^2 $ appears. 
Then since $R(\bm z) = \sum_{(j_1,j_2)\in\E}R_{j_1j_2}^\text{rot}(z_{j_1},z_{j_2}) - \nu$, $r_{\bm 0} = 4\sum_{(j_1,j_2)\in\E}\Omega_{j_1j_2}^2 - \nu$ as a real number.

Notice $R$ can be seen as the $n$-dimensional $z$-transform of the sequence $\{r_{\bm k}\}_{\bm k}$ over $\mathbb{T}^n$ according to \cite[Def. 2.1.1]{gregor1988multidimensional}.
Pick a scalar $\epsilon>0$ small enough such that $R(\bm z)>\epsilon>0$ for all $z \in \mathbb{T}^n$, then based on \cite[Thm. 2.1.5]{gregor1988multidimensional} we have 
\begin{align}
    r_{\bm 0} &= \frac{1}{(2\pi i)^n}\int_{\mathbb{T}^n} R(\bm z) z_1^{-1}z_2^{-1}\cdots z_n^{-1}d\bm z \\
    & > \frac{1}{(2\pi i)^n}\int_{\mathbb{T}^n} \epsilon z_1^{-1}z_2^{-1}\cdots z_n^{-1}d\bm z\\
    & = \frac{\epsilon}{(2\pi i)^n}\oint_{\mathbb{S}^1}\cdots\oint_{\mathbb{S}^1} z_1^{-1}\cdots z_n^{-1}dz_1\cdots dz_n\\
    & = \frac{\epsilon}{(2\pi i)^n} \cdot \left(\prod_{j=1}^n\oint_{\mathbb{S}^1} z_j^{-1}dz_j\right) \\
    & = \frac{\epsilon}{(2\pi i)^n} \cdot \left(\prod_{j=1}^n 2\pi i\right)\\
    & = \epsilon > 0.
\end{align}
\end{proof}

\begin{prop}
\label{rem:complex_conj_in_K}
The indices $\bm k \in \K$ of the non-zero coefficients of $R$ satisfy the following properties: 
\begin{enumerate}
    \item $\sum_{j=1}^nk_j = 0$;
    \item $-1\leq k_j\leq 1$ for all $j \in \{1,\cdots,n\}$;
    \item if $\bm k$ is not the 0 vector, then only 2 component of $\bm k$ are nonzero; 
    \item for each $\bm{k} \in \K$, $r_{\bm{k}} = r_{\bm{-k}}^\dagger$ where $^\dagger$ denotes the complex conjugate.
\end{enumerate}
\end{prop}
\begin{proof}
The statement follows directly from \eqref{eq:R_rot_j1j2} and the fact that $R(\bm z) = \sum_{(j_1,j_2)\in\E}R_{j_1j_2}^\text{rot}(z_{j_1},z_{j_2}) - \nu$.
\end{proof}

\subsection{Factorization and SDP}
In this subsection we factorize $R(\bm z)$ while preserving the sparsity structure as in Assumption \ref{assum: rip}.
For each $\ell\in\{1,\ldots,p\}$, let $\phi_\ell(\bm z)$ denote the column vector that stores the constant 1, every $z_{\texttt i}$, and every $z_{\texttt i_1}z_{\texttt i_2}$ where $\texttt i, \texttt i_1, \texttt i_2\in\I_\ell$ and $\texttt i_1\neq\texttt i_2$.
Therefore $\phi_\ell(\bm z)$ contains $\varrho_\ell:=1+|\I_\ell|+\binom{|\I_\ell|}{2}$ elements.
Let $\phi(\bm z):= [\phi_1^\top(\bm z), \phi_2^\top(\bm z),\cdots,\phi_p^\top(\bm z)]^\top$ which stacks all elements in $\phi_\ell(\bm z)$ for all $\ell$, thus it contains $\varrho:=\sum_{\ell=1}^p\varrho_\ell$ elements.
Let $\{\bm k_j\}_j$ be a sequence of $\bm k_j$ as the degree of $[\phi(\bm z)]_{(j,1)}$ in $\bm z$ for all $j=1,\cdots\varrho$, thus $\{-\bm k_j\}_j$ collects the orders of elements in $\phi^\dagger(\bm{z})$.
Notice that $\phi(\bm z)$ may contain repeated elements.
In addition, note $\Phi(\bm z):=\phi(\bm z) \cdot \phi^\dagger(\bm z)\in\C^{\varrho\times\varrho}$ contains all terms that appear in $R(\bm z)$, thus for the rest of this manuscript $\K$ is set to be the collection of degrees of all elements in $\Phi(\bm z)$ for clarity.

For arbitrary $\bm k\in\K$, define a binary matrix $\Theta_{\bm k}\in\R^{\varrho\times\varrho}$ that is zero everywhere except $[\Theta_{\bm k}]_{(j_1, j_2)}=1$ whenever $\bm k_{j_2} - \bm k_{j_1} = \bm k$, thus $tr(\Theta_{\bm k}\cdot \Phi(\bm z)) = \nnz(\Theta_{\bm k})\cdot\bm{z^k}$.
We then introduce the following 2 lemmas that can be used to check the positivity of $R$ using $\Theta_{\bm k}$ and $\Phi$.

\begin{lem}
\label{lem: rk match}
If there exists a matrix $Q\in\C^{\varrho\times \varrho}$ such that $r_{\bm k} = tr(\Theta_{\bm k}\cdot Q)$ holds for each $\bm k\in\K$, then $R(\bm z) = tr(\Phi(\bm z)\cdot Q)$.
\end{lem}
\begin{proof}
Notice $\Phi(\bm z) = \sum_{\bm k\in\K} \Theta_{\bm k}\bm{z^k}$, then 
\begin{align}
    R(\bm z) &= \sum_{\bm k\in\K} r_{\bm k} \bm{z^k} = \sum_{\bm k\in\K}tr(\Theta_{\bm k}\cdot Q)\bm{z^k}\\
    &= tr(\sum_{\bm k\in\K} \Theta_{\bm k}\bm{z^k}\cdot Q)= tr(\Phi(\bm z)\cdot Q)
\end{align}
in which the third equality comes from the facts that $\bm z^{\bm k}$ is a scalar and that $tr(\cdot)$ is a linear operator.
\end{proof}
\begin{lem} 
\label{lem: Q pos then f pos}
If $R(\bm z) = tr(\Phi(\bm z)\cdot Q)$ for a $Q\in\C^{\varrho\times \varrho}$ that is positive semi-definite, then $f(\bm{c,s})-\nu\geq 0$ over $\mathcal C$.
\end{lem}
\begin{proof}
If $Q\succeq 0$, then $Q=Y^*Y$ for some complex matrix $Y$ of proper size \cite[Corollary 7.2.9]{horn2012matrix}.
Therefore
\begin{align}
    R(\bm z) &= tr(\phi(\bm z)\phi^\dagger(\bm{z}) \cdot Y^*Y) \\
    &= tr(\phi^\dagger(\bm z)Y^\dagger Y\phi(\bm z))\\
    &=\phi^\dagger(\bm z)Y^\dagger Y\phi(\bm z)\\
    &= \|Y\phi(\bm z)\|_2^2
\end{align}
which is non-negative for all $\bm z$. 
Since $R(\bm z)$ is equivalent to $f(\bm{c,s})-\nu$ for $\bm{z} \in \mathbb{T}^n$ and $c_i^2 + s_i^2 = 1$ for all $i \in \{1,\ldots,n\}$, then $f(\bm{c,s})-\nu\geq 0$ where $c_i^2 + s_i^2 = 1$ for all $i \in \{1,\ldots,n\}$.
\end{proof}

One can determine whether a complex-valued matrix $Q$ is positive by checking the positivity of an equivalent real-valued matrix:
\begin{lem}[Eq. (6.26) in \cite{aps2018mosek}]
\label{lem: Q psd}
Let $Q = Q_r + i\cdot Q_i$ be a complex-valued matrix where $Q_r$ and $Q_i$ represent the real and imaginary parts of $Q$. If $Q$ is Hermitian, i.e. $Q_r$ is symmetric and $Q_i$ is skew-symmetric, then
\begin{equation}
    Q\succeq0 \iff \begin{bmatrix}
    Q_r & -Q_i \\ Q_i & Q_r
    \end{bmatrix}\succeq 0.
\end{equation}
\end{lem}

Given the two Lemmas presented above, we then seek to find a matrix $Q$ that is Hermitian and satisfies $r_{\bm k} = tr(\Theta_{\bm k}\cdot Q)$ for each $\bm k\in\K$.
Notice $Q$ being Hermitian is viable since $\bm{r_k} = \bm{r_{-k}}^*$ for all $\bm{k}\in\K$ as stated in Lemma \ref{rem:complex_conj_in_K}.
In addition, $Q$ is also expected to be block-diagonal as in \eqref{eq: Q blkdiag} in order to be compatible with the sparsity structure imposed by $\phi(\bm z)$: 
\begin{equation}
\label{eq: Q blkdiag}
    Q = \begin{bmatrix}
        Q_1  &  &  & \\
        & Q_2 &  & \\
        &  &  \ddots &\\
        &  &  & Q_p
    \end{bmatrix}\in\C^{\varrho\times\varrho}
\end{equation}
where $Q_\ell\in\C^{\varrho_\ell\times\varrho_\ell}$ for all $\ell\in\{1,\cdots,p\}$.

To find Q and the optimal $\nu$, we can solve the following optimization
\begin{align*}
    \sup_{Q,\nu}  \quad & \nu   \hspace{3cm}\texttt{(Opt)}\\
    \text{s.t.} \quad & tr(\Theta_{\bm k}\cdot Q) = r_{\bm k}, ~ \forall \bm k\in\K\\
    & Q \text{ is Hermitian, } Q\succeq 0
\end{align*}
where $\lambda\in\R$ is a slack variable.
To formulate this problem over real-valued matrices, define a symmetric block matrix 
\begin{equation}
\label{eq: defn X}
    X = \begin{bmatrix} 
    Q_r  & -Q_i\\
    Q_i & Q_r 
    \end{bmatrix},
\end{equation}
then \texttt{(Opt)} can be written as an equivalent Semi-Definite Program (SDP) as in \cite[eq. (1.1.1)]{wolkowicz2012handbook}:
\begin{align*}
    \inf_{X,\nu}  \quad & -\nu   \hspace{4cm}\texttt{(P)}\\
    \text{s.t.} \quad & tr(A_m\cdot X) = b_m, ~ \forall m = 1,2,\cdots,M\\
    &X\succeq 0\\
    &X\in \mathbb S_{2\varrho}
\end{align*}
in which $X\succeq 0$ ensures that $Q\succeq 0$ based on Lemma \ref{lem: Q psd}.
The cost function of \texttt{(P)} actually tries to maximize $\nu$, and the equality constraints in \texttt{(P)} enforce \eqref{eq: Q blkdiag}, \eqref{eq: defn X}, and $r_{\bm k} = tr(\Theta_{\bm k}\cdot Q)$ for all $\bm k\in\K$.
In particular, the equality constraints in \texttt{(P)} are enforcing:
\begin{enumerate}

    \item \textbf{Block Diagonal}:
    To ensure $Q$, or equivalently $Q_r$ and $Q_i$, are block diagonal as in \eqref{eq: Q blkdiag}, define
    \begin{equation}
        Q^\text{BD} = \begin{bmatrix}
        1_{\varrho_1} & & &\\
        & 1_{\varrho_2} & &\\
        & & \ddots &\\
        & & & 1_{\varrho_p}
        \end{bmatrix},
    \end{equation}
    $\Gamma_1 := \{(\gamma_1,\gamma_2)\in\Z_+\mid[Q^\text{BD}]_{(\gamma_1\gamma_2)}=0,1\leq\gamma_1<\gamma_2\leq\varrho\}$, $\Gamma_2 := \{(\gamma_1,\gamma_2+\varrho)\in\Z_+\mid[Q^\text{BD}]_{(\gamma_1\gamma_2)}=0,1\leq\gamma_1<\gamma_2\leq\varrho\}$, and $\Gamma:=\Gamma_1\cup\Gamma_2$.
    For each $\gamma\in\Gamma$, define $b_\gamma^\text{BD} = 0$ and $A_\gamma^\text{BD}\in\R^{2\varrho\times2\varrho}$ that is 0 anywhere except $[A_{\gamma}^\text{BD}]_{\gamma}=1$.
    Then given skew symmetric $Q_i$, $Q_r$ and $Q_i$ are both block diagonal if $tr(\sym(A_\gamma^\text{BD})\cdot X) = b_\gamma^\text{BD}$ holds for all $\gamma\in\Gamma$.
    Since $Q^\text{BD}$ is symmetric, it is easy to check $|\Gamma_1| = |\Gamma_2| = 0.5(\varrho^2 - \sum_{\ell=1}^p\varrho_\ell^2)$ and $\Gamma_1\cap\Gamma_2 = \emptyset$.
    Therefore in total we have $\varrho^2 - \sum_{\ell=1}^p\varrho_\ell^2$ such constraints.

    
    \item \textbf{Block Matching}: In $X$ we need its diagonal blocks appearing as $Q_r$ in \eqref{eq: defn X} to be identical to one another.
    This condition can be enforced by defining $b_{\alpha_1\alpha_2}^\text{BM} = 0$ and $A_{\alpha_1\alpha_2}^\text{BM}\in\R^{2\varrho\times2\varrho}$ which is 0 everywhere except 
    \begin{equation}
        [A_{\alpha_1\alpha_2}^\text{BM}]_{(\alpha_1, \alpha_2)} = -[A_{\alpha_1\alpha_2}^\text{BM}]_{(\alpha_1+\varrho, \alpha_2+\varrho)} = 1
    \end{equation}
    for all $\alpha_1,\alpha_2\in\{1,2,\cdots,\varrho\}$.
    Therefore $tr(\sym(A_{\alpha_1\alpha_2}^\text{BM})\cdot X) = b_{\alpha_1\alpha_2}^\text{BM}$ guarantees that the $(\alpha_1\alpha_2)$-th elements of the first and second diagonal blocks in $X$ are the same. 
    Because $X$ is symmetric, we have $1+2+\cdots+\varrho = 0.5(\varrho^2+\varrho)$ such constraints.
    
    \item \textbf{Skew Symmetry}: 
    To ensure $[Q_i]_{(\beta_1,\beta_2)} = -[Q_i]_{(\beta_2,\beta_1)}$ for all indices $\beta_1,\beta_2\in\{1,2,\cdots,\varrho\}$, we define the following constraints:
    \begin{equation}
        tr(\sym(A_{\beta_1\beta_2}^\text{SS})\cdot X) = b_{\beta_1\beta_2}^\text{SS}
    \end{equation}
    where $b_{\beta_1\beta_2}^\text{SS}=0$ and $A_{\beta_1\beta_2}^\text{SS}\in\R^{2\varrho\times2\varrho}$ is 0 everywhere except
    \begin{itemize}
        \item $[A_{\beta_1\beta_2}^\text{SS}]_{(\beta_1,\beta_2+\varrho)} = 1$ if $\beta_1=\beta_2$;
        \item $[A_{\beta_1\beta_2}^\text{SS}]_{(\beta_1,\beta_2+\varrho)} = [A_{\beta_1\beta_2}^\text{SS}]_{(\beta_2,\beta_1+\varrho)} = 1$ if $\beta_1\neq\beta_2$.
    \end{itemize}
    The number of such constraints is $\varrho + \big(1+2+\cdots+(\varrho-1)\big) = 0.5(\varrho^2+\varrho)$,  where the first $\varrho$ comes from the cases when $\beta_1=\beta_2$ and the remainder of the summation comes from cases when $\beta_1\neq\beta_2$. 
    
    \item \textbf{Term Matching}: To ensure $r_{\bm k} = tr(\Theta_{\bm k}\cdot Q)$ for all $\bm k\in\K$, enforce the following constraints:
    \begin{align}
        & tr(\sym(A_{\bm k}^\text{TMre})\cdot X) = b_{\bm k}^\text{TMre}\\  
        & tr(\sym(A_{\bm k}^\text{TMim})\cdot X) = b_{\bm k}^\text{TMim}
    \end{align}
    where
    \begin{align}
        A_{\bm k}^\text{TMre} = \begin{bmatrix}
         \Theta_{\bm k} & \\ & 0_{\varrho}
        \end{bmatrix}, & ~ b_{\bm k}^\text{TMre} = \text{real}(r_{\bm k}),\\
        A_{\bm k}^\text{TMim} = \begin{bmatrix}
         0_{\varrho} &\Theta_{\bm k}  \\ & 0_{\varrho}
        \end{bmatrix}, & ~ b_{\bm k}^\text{TMim} = \text{im}(r_{\bm k}).
    \end{align}
    In total, there are $2|\K|$ such constraints.

\end{enumerate}

Notice the second and third categories of constraints ensure that $X$ has the structure of \eqref{eq: defn X}.
Let $\{b_m\}_m$ be the collection of all possible $b_{\gamma}^\text{BD}$, $b_{\alpha_1\alpha_2}^\text{BM}$, $b_{\beta_1\beta_2}^\text{SS}$, $b_{\bm k}^\text{TMre}$ and $b_{\bm k}^\text{TMim}$;  
let $\{A_m\}_m$ be the collection of all possible $\sym(A_{\gamma}^\text{BD})$, $\sym(A_{\alpha_1\alpha_2}^\text{BM})$, $\sym(A_{\beta_1\beta_2}^\text{SS})$, $\sym(A_{\bm k}^\text{TMre})$ and $\sym(A_{\bm k}^\text{TMim})$. 
Then in total \texttt{(P)} has $M = 2\varrho^2 + \varrho - \sum_{\ell=1}^p\varrho_\ell^2+ 2|\K|$ equality constraints. 

\subsection{Sparse SDP}

This section describes how to formulate \texttt{(P)} as a smaller semidefinite program by utilizing the sparsity of the problem formulation. 
We refer to this smaller semidefinite program as the Complex-domain SDP (CSDP). 

Notice $X\in\mathbb S_{2\varrho}$ in \texttt{(P)} is indeed sparse when $p$ is large, or equivalently each $\varrho_\ell$ is much smaller than $\varrho$.
This can be seen by comparing the size of $Q$ against the sizes of all $Q_\ell$ in \eqref{eq: Q blkdiag}.
In other words, $\varrho^2 = (\sum_{\ell=1}^p\varrho_\ell)^2$ is much larger than $\sum_{\ell=1}^p\varrho_\ell^2$ when $p$ is large.
We then can simplify \texttt{(P)} into an SDP with smaller size by transferring $X$ into a block diagonal matrix using the block diagonal structure in $Q$.

To block diagonalize $X$, for arbitrary $\varepsilon:=(\varepsilon_1,\varepsilon_2)\in\{1,2,\cdots,2\varrho\}^2$, denote $E_{\varepsilon}$ the matrix generated by exchanging the $\varepsilon_1$-th and $\varepsilon_2$-th columns of $I_{2\varrho}$.
Notice $E_{\varepsilon}E_{\varepsilon}^\top = E_{\varepsilon}^\top E_{\varepsilon}= I_{2\varrho}$, and $E_{\varepsilon} = I_{2\varrho}$ if $\varepsilon_1 = \varepsilon_2$.
Define a linear operator $\mE_\varepsilon:\mathbb S_{2\varrho}\rightarrow \mathbb S_{2\varrho}$ as
\begin{equation}
    \mE_\varepsilon(A) = E_\varepsilon^\top A E_\varepsilon
\end{equation}
that switch the $\varepsilon_1$-th row and column with the $\varepsilon_2$-th row and column of $A$.
Then for an subset $\mathcal E:= \{\varepsilon^\mu\}_\mu\subset\{1,2,\cdots,2\varrho\}^2$, define
\begin{equation}
    \mE_{\mathcal E}:= \mE_{\varepsilon^{|\mathcal E|}} \circ \mE_{\varepsilon^{|\mathcal E|-1}} \circ \cdots \circ \mE_{\varepsilon^{1}},
\end{equation}
and the following lemma holds.
\begin{lem}
\label{lem: e-val is not changed}
$\mE_{\mathcal E}(A)$ shares the same eigenvalues with $A$ for arbitrary matrix $A\in\mathbb S_{2\varrho}$ and arbitrary sequence $\mathcal E\subset\{1,2,\cdots,2\varrho\}^2$.
\end{lem}
\begin{proof}
Due to \cite[Corollary 3.3.1]{Serre2002MatricesTA}, symmetric matrix $A$ can be diagonalized as $A=V^\top\Lambda V$ where $V$ is orthogonal, i.e. $VV^\top=V^\top V = I_{2\varrho}$ and $\Lambda$ is a diagonal matrix. Notice the diagonal of $\Lambda$ collects all eigenvalues of $A$.
Then for arbitrary $\varepsilon\in\{1,2,\cdots,2\varrho\}^2$, 
\begin{equation}
\mE_\varepsilon(A) = E_\varepsilon^\top AE_\varepsilon=E_\varepsilon^\top V^\top\Lambda V E_\varepsilon = (VE_\varepsilon)^\top\Lambda (VE_\varepsilon),   
\end{equation}
thus $\mE_\varepsilon(A)$ shares the same eigenvalues with $A$.
The claim then follows by iterativly applying the above computation $|\mathcal E|$ times.
\end{proof}

Before presenting a lemma and a corollary that can be useful block diagonalze $X$, we introduce one more notation.
Given an arbitrary matrix $A\in\mathbb S_{2\varrho}$ whose diagonal elements are all elements in $\{1,2,\cdots,p\}$, define
\begin{equation}
    \mathbb L_\ell(A) := \begin{bmatrix}
[A]_{(j_1,j_1)} & [A]_{(j_1,j_2)} & \cdots & [A]_{(j_1,j_l)}\\
[A]_{(j_2,j_1)} & [A]_{(j_2,j_2)} & \cdots & [A]_{(j_2,j_l)}\\
\vdots & \vdots & \vdots & \vdots \\ 
[A]_{(j_l,j_1)} & [A]_{(j_l,j_2)} & \cdots & [A]_{(j_l,j_l)}
\end{bmatrix}
\end{equation}
where $\{j_1,j_2,\cdots,j_l\}$ is a collection of all possible values of index $j_k$ such that $[A]_{(j_k,j_k)}=\ell$, and we assume $1\leq j_1<j_2<\cdots<j_l$.

\begin{lem}
\label{lem: cnst blk over epsilon}
    For any $\ell\in\{1,2,\cdots,p\}$ and matrix $A\in\mathbb S_{2\varrho}$, if $\mathbb L_\ell(A) = \ell_{\varrho_\ell}$, then $\mathbb L_\ell(\mE_\varepsilon(A)) = \ell_{\varrho_\ell}$ for arbitrary $\varepsilon=(\varepsilon_1,\varepsilon_2)\in\{1,2,\cdots,2\varrho\}^2$. 
\end{lem}
\begin{proof}
    Rewrite $\mE_\varepsilon(A)$ as $E_\varepsilon^\top \bar A$ where $\bar A = AE_\varepsilon$, then $\mE_\varepsilon(A)$ can be computed as first switch the $\varepsilon_1$-th and $\varepsilon_2$-th columns of $A$, and secondly switch the $\varepsilon_1$-th and $\varepsilon_2$-th rows of $\bar A$.
    Therefore the $\varepsilon_1$-th column of $\mE(A)$ is the same as the $\varepsilon_2$-th column of $A$ except $[\mE_\varepsilon(A)]_{(\varepsilon_1,\varepsilon_1)} = [A]_{(\varepsilon_2,\varepsilon_2)}$ and $[\mE_\varepsilon(A)]_{(\varepsilon_1,\varepsilon_2)} = [A]_{(\varepsilon_2,\varepsilon_1)}$.
    Similarly the $\varepsilon_2$-th column of $\mE(A)$ is the same as the $\varepsilon_1$-th column of $A$ except $[\mE_\varepsilon(A)]_{(\varepsilon_2,\varepsilon_1)} = [A]_{(\varepsilon_1,\varepsilon_2)}$ and $[\mE_\varepsilon(A)]_{(\varepsilon_2,\varepsilon_2)} = [A]_{(\varepsilon_1,\varepsilon_1)}$.
    Since $\mE(A)$ is symmetric, discussion on rows of $\mE(A)$ is omitted. 
    
    Denote $\mathcal L_\ell = \{j\in\Z\mid 1\leq j\leq 2\varrho, [A]_{(j,j)}=\ell\}$.
    If $\{\varepsilon_1,\varepsilon_2\}\cap\mathcal L_\ell=\emptyset$, then trivially $\mathbb L_\ell(\mE_\varepsilon(A)) = \ell_{\varrho_\ell}$ since $\mathbb L_\ell(A) = \mathbb L_\ell(\mE_\varepsilon(A))$.
    If any of $\varepsilon_1$ or $\varepsilon_2$ belongs to $\mathcal L_\ell$, say $[A]_{(\varepsilon_1,\varepsilon_1)}=\ell$ for example, then $[\mE_\varepsilon(A)]_{(\varepsilon_2,\varepsilon_2)} = [\mE_\varepsilon(A)]_{(j,\varepsilon_2)} = \ell$ for all $j\in\mathcal L_\ell\setminus\{\varepsilon_1\}$ given $\mathbb L_\ell(A)= \ell_{\varrho_\ell}$.
    Therefore $\mathbb L_\ell(\mE_\varepsilon(A)) = \ell_{\varrho_\ell}$.

\end{proof}

\begin{cor}[from Lemma \ref{lem: cnst blk over epsilon}]
\label{cor: cnst blk over Epsilon}
    For any $\ell\in\{1,2,\cdots,p\}$ and matrix $A\in\mathbb S_{2\varrho}$, if $\mathbb L_\ell(A) = \ell_{\varrho_\ell}$, then $\mathbb L_\ell(\mE_{\mathcal E}(A)) = \ell_{\varrho_\ell}$ for arbitrary $\mathcal E\subset\{1,2,\cdots,2\varrho\}^2$.
\end{cor}
\begin{proof}
The claim can be shown by applying Lemma \ref{lem: cnst blk over epsilon} $|\mathcal E|$ times. 
\end{proof}

We now present the key theorem that block diagonalize $X$.
\begin{thm}
\label{thm: alg 1}
There exists a sequence $\mathcal E:= \{\varepsilon^\mu\}_\mu\subset\{1,2,\cdots,2\varrho\}^2$ such that $\mE_{\mathcal E}(X)$ is a block diagonal matrix.
\end{thm}
\begin{proof}
To block diagonalize $X$, we mask $X$ by a matrix 
\begin{equation}
    B = \begin{bmatrix}
    1_{\varrho_1} & & & &1_{\varrho_1} & & &\\
    & 2_{\varrho_2}& & & &2_{\varrho_2} & &\\
    & & \ddots & & & &\ddots &\\
    & & & p_{\varrho_p} & & & & p_{\varrho_p}\\
    1_{\varrho_1} & & & &1_{\varrho_1} & & &\\
    & 2_{\varrho_2}& & & &2_{\varrho_2} & &\\
    & & \ddots & & & &\ddots &\\
    & & & p_{\varrho_p} & & & & p_{\varrho_p}
    \end{bmatrix}
\end{equation}
that has the same size and block structure as $X$. 
Notice that in $B$ we assign the entire blocks that correspond to the real and imagine portions of $Q_\ell$ in $X$ by the constant $\ell$, and that all nonzero elements on the $j$-th column and row of $B$ have the same value for arbitrary $j=1,\cdots,2\varrho$. 
Our goal to create a sequence $\mathcal E$ such that
\begin{equation}
\label{eq: desired B}
    \mE_{\mathcal E}(B) = \begin{bmatrix}
    1_{\varrho_1} & 1_{\varrho_1} & & & & &\\
    1_{\varrho_1} & 1_{\varrho_1} & & & & &\\
    & & 2_{\varrho_2} & 2_{\varrho_2} & & &\\
    & & 2_{\varrho_2} & 2_{\varrho_2} & & &\\
    & & & &\ddots & &\\
    & & & & & p_{\varrho_p} & p_{\varrho_p}\\
    & & & & & p_{\varrho_p} & p_{\varrho_p}
    \end{bmatrix},
\end{equation}
then $\mE_{\mathcal E}(X)$ is a block diagonal matrix as well with the same sizes of blocks in $\mE_{\mathcal E}(B)$.

Notice $B\in\mathbb S_{2\varrho}$ and $\mathbb L_\ell(B) = \ell_{\varrho_\ell}$ for all $\ell=1,2,\cdots,p$, then it suffices to show the existence of $\mathcal E$ that rearranges $\diag(B)$ into
\begin{equation}
\label{eq: desired diagB}
    \tilde B:=\begin{bmatrix}
    1_{1\times2\varrho_1} & 2_{1\times2\varrho_2} & \cdots & p_{1\times2\varrho_p} 
    \end{bmatrix}
\end{equation}
according to Corollary \ref{cor: cnst blk over Epsilon}.
We show the existence of such $\mathcal E$ by an inductive argument. 

We start by $\mathcal E$ as an empty set and $\diag(B)$ as
\begin{equation}
    \begin{bmatrix}
    1_{1\times\varrho_1} & 2_{1\times\varrho_2} & \cdots & p_{1\times\varrho_p} & 1_{1\times\varrho_1} & 2_{1\times\varrho_2} & \cdots & p_{1\times\varrho_p}
    \end{bmatrix}
\end{equation}
which contains $2\varrho$ elements.
Notice numbers with value 1 in $\diag(B)$ are not gathered as in \eqref{eq: desired diagB}, but the first $\varrho_1$ elements in $\diag(B)$ are already the same as the first $\varrho_1$ elements in $\tilde B$. 
We can then enlarge set $\mathcal E$ by a new element $\varepsilon^1 = (\varrho_1+1, \varrho+1)$.
Due to the proof of Lemma \ref{lem: cnst blk over epsilon}, $\varepsilon^1$ bring the first number with value 1 among the last $2\varrho-\varrho_1$ elements of $\diag(B)$ next to the first $\varrho_1$ elements of $\diag(B)$ on the right, so that $\diag(\mE_{\mathcal{E}\cup\{\varepsilon^1\}}(B))$ and $\tilde B$ share the same first $\varrho_1+1$ elements.

Now suppose there exists some set $\mathcal E\subset\{1,2,\cdots,2\varrho\}^2$ such that $\diag(\mE_\mathcal{E}(B))$ and $\tilde B$ accord with the first $k$ elements, but not the $(k+1)$-th element.
Then we can enlarge $\mathcal E$ by a new element $\varepsilon^\mu=(\varepsilon_1,\varepsilon_2)$ where $\varepsilon_1 = k+1$ and
\begin{equation}
    \varepsilon_2 = \min\{j\in\Z\mid j>\varepsilon_1, [\mE_\mathcal{E}(B)]_{(j,j)} = [\tilde B]_{(1,\varepsilon_1)}\}.
\end{equation}
Notice $\varepsilon_2$ is not empty since $\diag(\mE_\mathcal{E}(B))$ and $\tilde B$ are composed of the same elements but sorted in different orders.
Then $\diag(\mE_{\mathcal{E}\cup\{\varepsilon^\mu\}}(B))$ and $\tilde B$ share the same first $k+1$ elements.

Therefore by induction, $\mathcal E$ can be expanded from empty set until $\diag(\mE_\mathcal{E}(B)) = \tilde B$.

\end{proof}

The proof of Theorem \ref{thm: alg 1} induces steps of constructing $\mathcal E$ as shown in Algorithm \ref{alg: compute E}.


\begin{algorithm}[h]
    \caption{Construction of $\mathcal E$}
    \label{alg: compute E}
    \begin{algorithmic}
        \REQUIRE $B$, $\tilde B$
        \STATE $\mathcal E\gets\emptyset$
        \STATE $\tilde B\gets\diag(B)$, $\varepsilon_1\gets 1$, $\varrho\gets0.5\cdot(\text{length of }\tilde B)$
        \FOR{$\varepsilon_1 = 1,2,\cdots,2\varrho$}
            \IF{$[B]_{(\varepsilon_1,\varepsilon_1)} \neq [\tilde B]_{(1,\varepsilon_1)}$}
                \STATE $\varepsilon_2\gets\min\{j\in\Z\mid j>\varepsilon_1, [\mE_\mathcal{E}(B)]_{(j,j)}  = [\tilde B]_{(1,\varepsilon_1)}  \}$
                \STATE $\varepsilon^\mu \gets (\varepsilon_1,\varepsilon_2)$, $\mathcal E \gets \mathcal E \cup \{\varepsilon^\mu\}$
            \ENDIF
        \ENDFOR
        \RETURN $\mathcal E$
    \end{algorithmic}
\end{algorithm}

Now consider the following sparse version of \texttt{(P)}
\begin{align*}
    \inf_{\nu,X_1,\cdots X_p}  \quad & -\nu   \hspace{5.5cm}\texttt{(SP)}\\
    \text{s.t.} \quad \quad ~
    & tr(\mE_{\mathcal E}(A_m)\cdot X') = b_m, ~ \forall m = 1,2,\cdots,M\\
    & X' = \begin{bmatrix}
    X_1 & & &\\ & X_2 & &\\ & & \ddots & \\& & & X_p
    \end{bmatrix}\\
    & X_\ell\in\mathbb S^{2\varrho_\ell}, ~ X_\ell\succeq 0, ~ \forall \ell = 1,2,\cdots,p
\end{align*}
which is easier to solve than \texttt{(P)} because instead of requiring a large size matrix $X$ being positive semi-definite as in \texttt{(P)}, matrices enforced to be positive semi-definite in \texttt{(SP)} are of much smaller sizes.
\begin{thm}
\label{thm: sp is the same as P}
The optimal solution $\nu'$ of \texttt{(SP)} accords with the optimal solution $\nu^*$ of \texttt{(P)}.
\end{thm}
\begin{proof}
Since a block diagonal matrix is positive semi-definite if and only if each of its blocks is positive semi-definite, then $X'\succeq 0$ given $X_\ell\succeq 0$ for all $\ell$.
Let $X = \mE_{\mathcal E}^{-1}(X')$, thus $X$ share the same eigenvalues with $X'$ and $X\succeq 0$.
Notice 
\begin{align}
    tr(A_m\cdot X) &= tr(A_m\cdot E_{\varepsilon^1}E_{\varepsilon^2}\cdots E_{\varepsilon^{|\mathcal E|}}X'E_{\varepsilon^{|\mathcal E|}}^\top\cdots E_{\varepsilon^2}^\top E_{\varepsilon^1}^\top)\\
    &= tr(E_{\varepsilon^{|\mathcal E|}}^\top\cdots E_{\varepsilon^2}^\top E_{\varepsilon^1}^\top A_m E_{\varepsilon^1}E_{\varepsilon^2}\cdots E_{\varepsilon^{|\mathcal E|}}\cdot X')\\
    &= tr( \mE_{\mathcal E}(A_m) \cdot X' ),
\end{align}
therefore \texttt{(SP)} and \texttt{(P)} have the same cost and constraints, and the claim follows. 
\end{proof}

\begin{remark}
Similar as the SBSOS formulation, the sparse SDP formulation can also be constructed and solved in hierarchy based on the maximal degree of elements in $\phi_\ell(\bm z)$ for all $\ell$.
In the above discussion, $\phi_\ell(\bm z)$ is defined to contain terms of degree no greater than 2, thus \texttt{(SP)} can be seen as the second step of the sparse SDP hierarchy.
One can eliminate all degree 2 terms in $\phi_\ell(\bm z)$ to get the first step of the sparse SDP hierarchy, or enlarge $\phi_\ell(\bm z)$ by higher degree terms to achieve higher hierarchy step.
\end{remark}

Finally we point out that constraints on the category of `Block Diagonal' in \texttt{(P)} are no longer necessary in \texttt{(SP)} since these constraints make $\mE_\varepsilon(X)$ block diagonal and $X' = \mE_\varepsilon(X)$ are built as a block diagonal matrix in \texttt{(SP)}.
Therefore we can further simplify \texttt{(SP)} by eliminating constraints on the category of `Block Diagonal' without influencing the solution. 
\section{Experimental Proof of Concept}

This section illustrates the performance of the pair of hierarchies.
First, we show the performance of the SBSOS-SLAM hierarchy at the first step of the hierarchy on a variety of state of the art datasets.
Second, we contrast the performance of the pair of hierarchies at several steps within the hierarchies on randomly generated fully connected Pose Graphs and one dataset.

\subsection{SBSOS-SLAM at the first hierarchy}

We evaluated the proposed \ac{SLAM} algorithm on the CityTrees10000 \cite{kaess2008isam} and Manhattan3500 \cite{olson2006fast} datasets by breaking the problem into sequences of 100 nodes and solving those graphs for which Mosek \cite{mosek_c_api} did not run into numerical instabilities.
For each sequence (taken individually), we used the proposed SBSOS-SLAM methodology to find the
optimal solution at the first hierarchy to the respective \ac{MLE} problem defined in \eqref{eq:pose_slam_poly_opt} and
\eqref{eq:landmark_slam_poly_opt}. For comparison, we also initialized Levenberg-Marquardt with a random initialization.

The median solve time for SBSOS-SLAM was 20.5507 seconds for the Manhattan3500 dataset and 161.8387 sec for the CityTree10000 dataset compared to less than a second on average for Levenberg-Marquardt. 
However, since our current implementation is based in Matlab and we are not attempting to satisfy the running intersection property optimally in these initial experiments, we believe there are a variety of extensions that can be made to improve scalability. 

\begin{figure}[tb]%
  \centering%
  \subfloat[Ground Truth - No Noise]{%
    \includegraphics[trim={4.1cm, 9cm, 4.5cm, 9.5cm},clip,width=0.5\columnwidth]{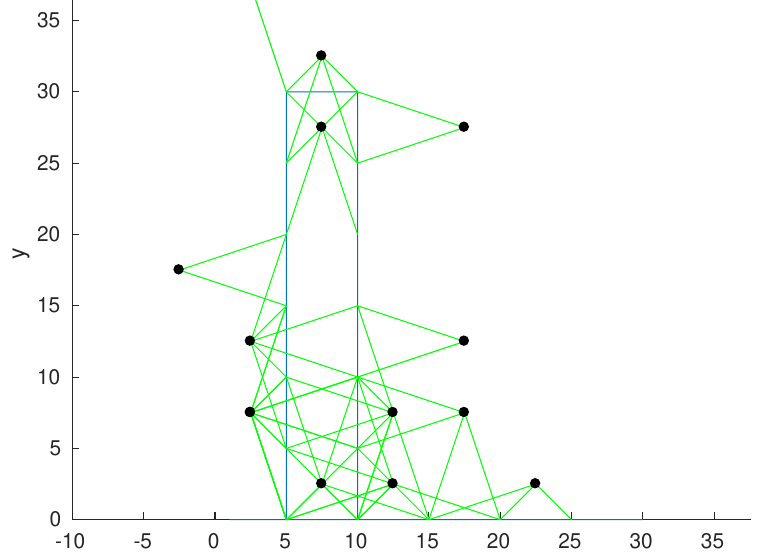}%
    \label{fig:CT100_gt}%
  }%
  \subfloat[Random Initialization]{%
    \includegraphics[trim={4.1cm, 9cm, 4.5cm, 9.5cm},clip,width=0.5\columnwidth]{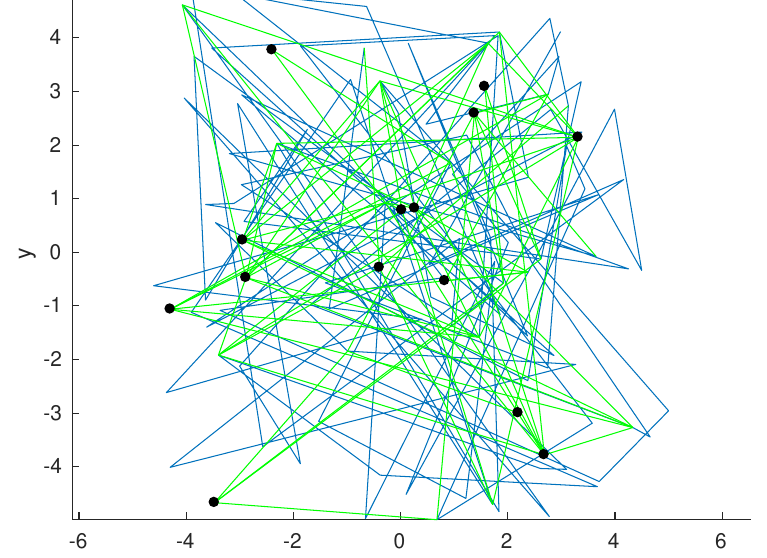}%
    \label{fig:CT100_random_init}%
  }%
  \hfil%
  \subfloat[Levenberg-Marquardt]{%
    \includegraphics[trim={4.1cm, 9cm, 4.5cm, 9.5cm},clip,width=0.5\columnwidth]{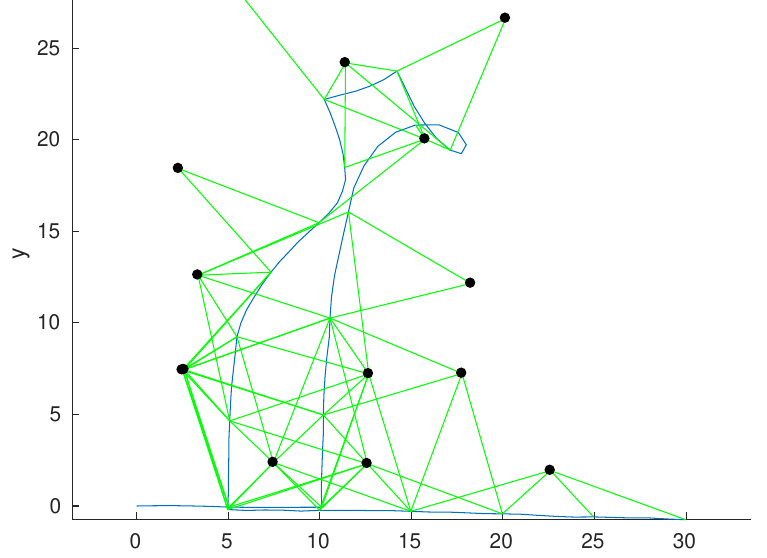}%
    \label{fig:CT100_LM}%
  }%
  \subfloat[SBSOS-SLAM]{%
    \includegraphics[trim={4.1cm, 9cm, 4.5cm, 9cm},clip,width=0.5\columnwidth]{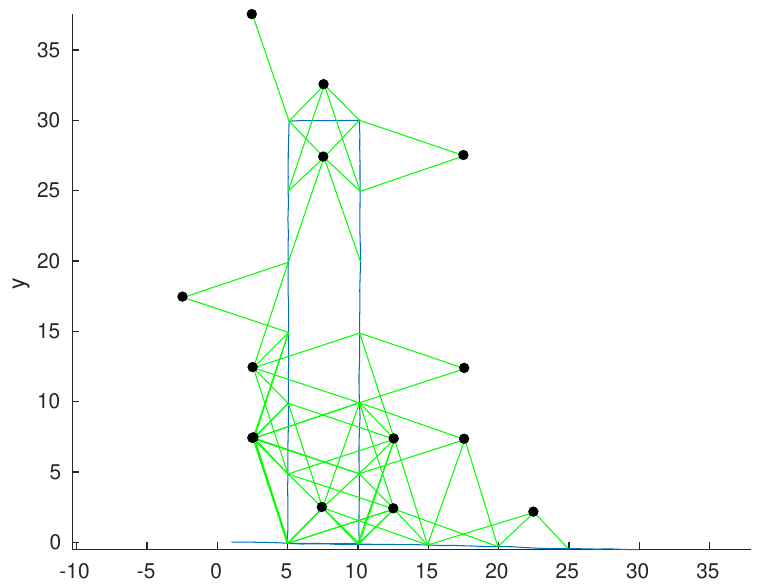}%
    \label{fig:CT100_SBSOS}%
  }%
  \caption{\small Sample estimated Landmark SLAM solution for 100 nodes of CityTrees10000 dataset \cite{kaess2008isam}.}
  \label{fig:cityTrees100}
\end{figure}

\begin{figure}[tb]%
  \centering%
  \subfloat[Ground Truth - No Noise]{%
    \includegraphics[trim={4.1cm, 9cm, 4.5cm, 9.5cm},clip,width=0.5\columnwidth]{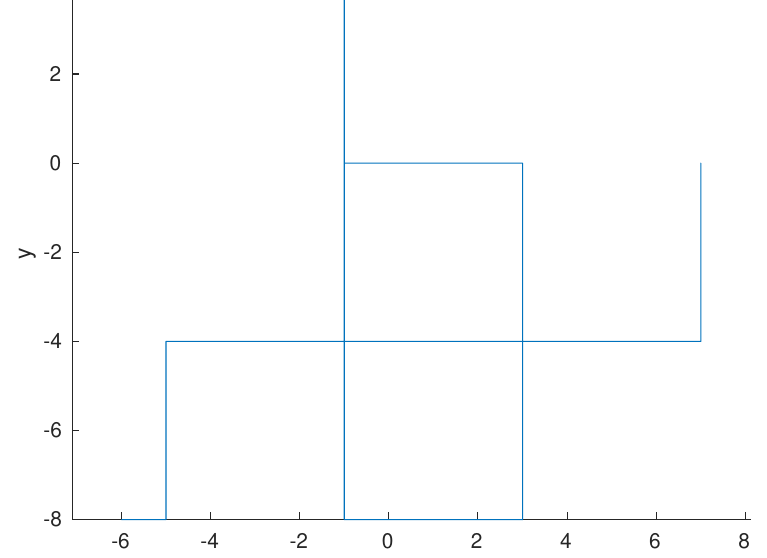}%
    \label{fig:man_gt}%
  }%
  \subfloat[Random Initialization]{%
    \includegraphics[trim={4.1cm, 9cm, 4.5cm, 9.5cm},clip,width=0.5\columnwidth]{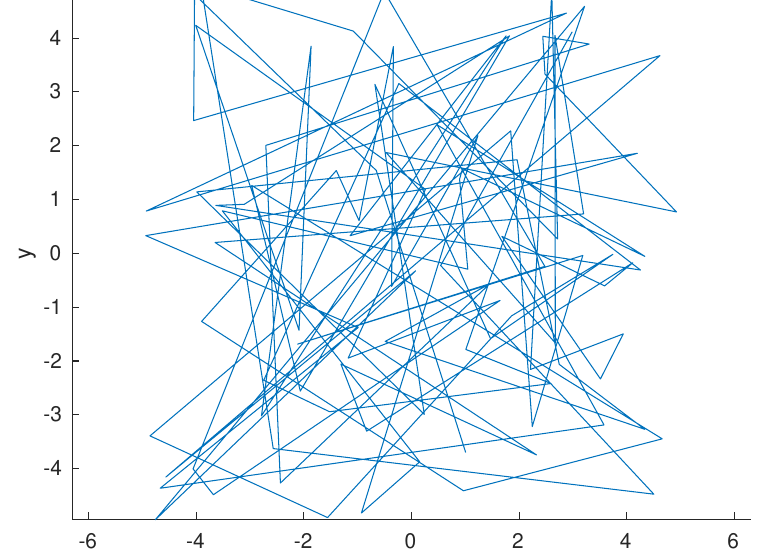}%
    \label{fig:man_random_init}%
  }%
  \hfil%
  \subfloat[Levenberg-Marquardt]{%
    \includegraphics[trim={4.1cm, 9cm, 4.5cm, 9.5cm},clip,width=0.5\columnwidth]{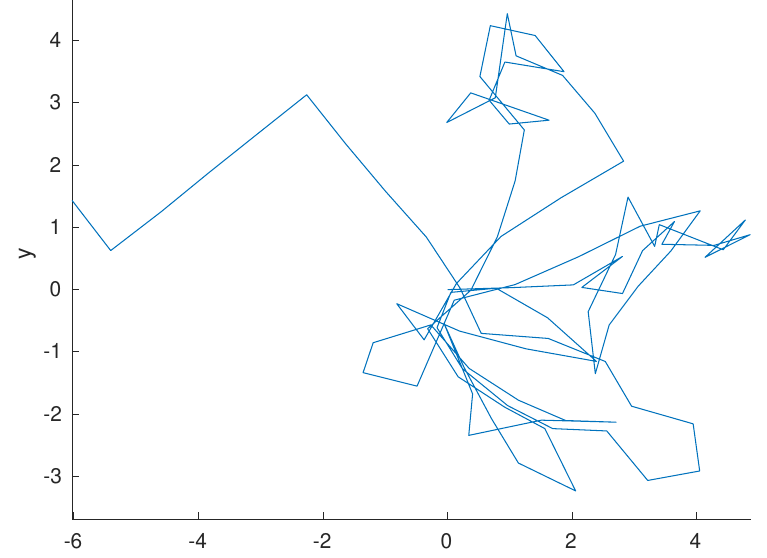}%
    \label{fig:man_LM}%
  }%
  \subfloat[SBSOS-SLAM]{%
    \includegraphics[trim={4.1cm, 9cm, 4.5cm, 9cm},clip,width=0.5\columnwidth]{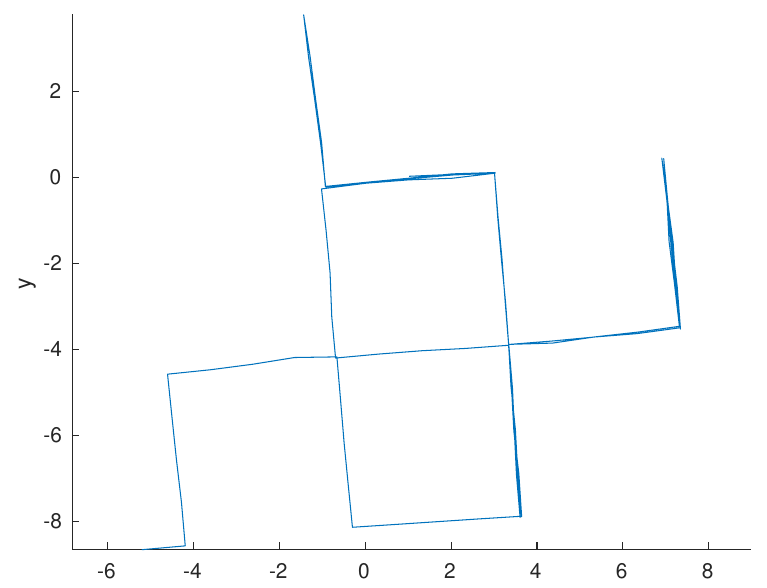}%
    \label{fig:man_SBSOS}%
  }%
  \caption{\small Sample estimated Pose Graph SLAM solution for 100 nodes of Manhattan3500 dataset \cite{olson2006fast}.}
  \label{fig:man3500}
\end{figure}

\begin{figure}[tb]%
  \centering%
  \includegraphics[trim={1cm, 6cm, 1cm, 6cm},clip,width=0.5\columnwidth]{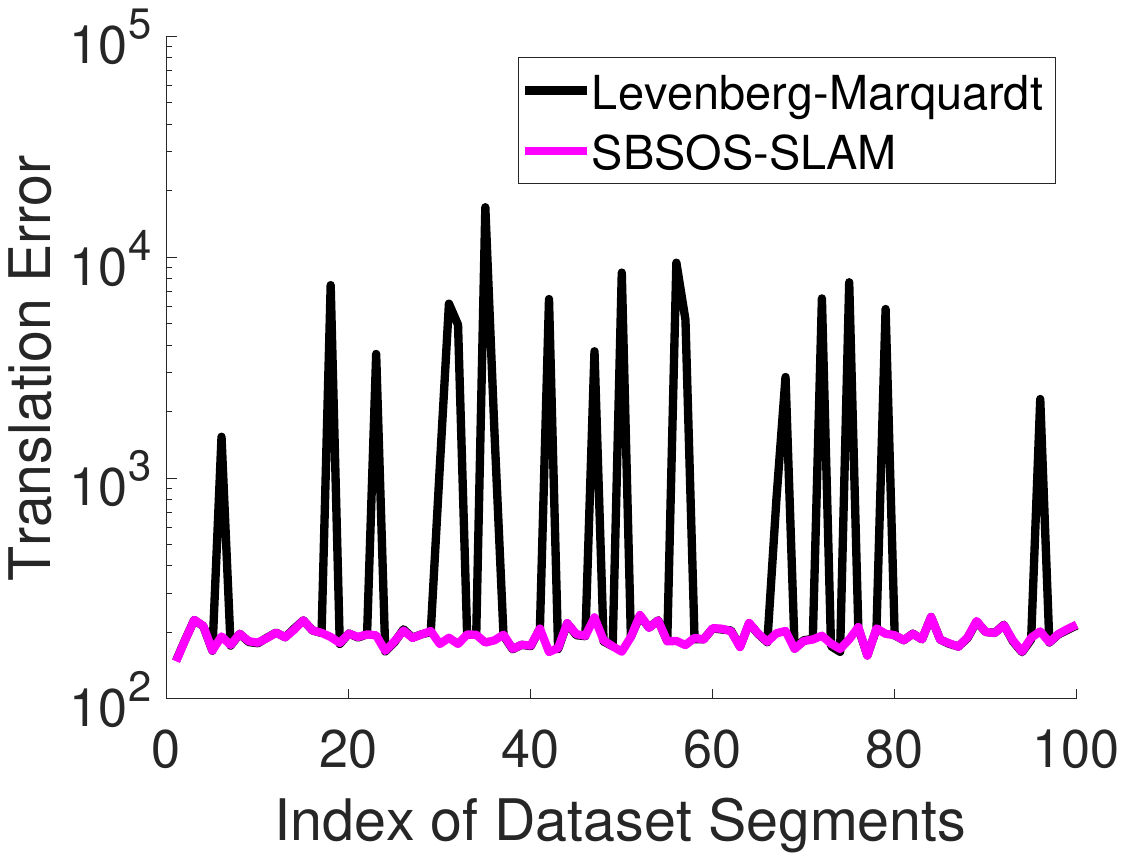}%
  \includegraphics[trim={1cm, 6cm, 1cm, 6cm},clip,width=0.5\columnwidth]{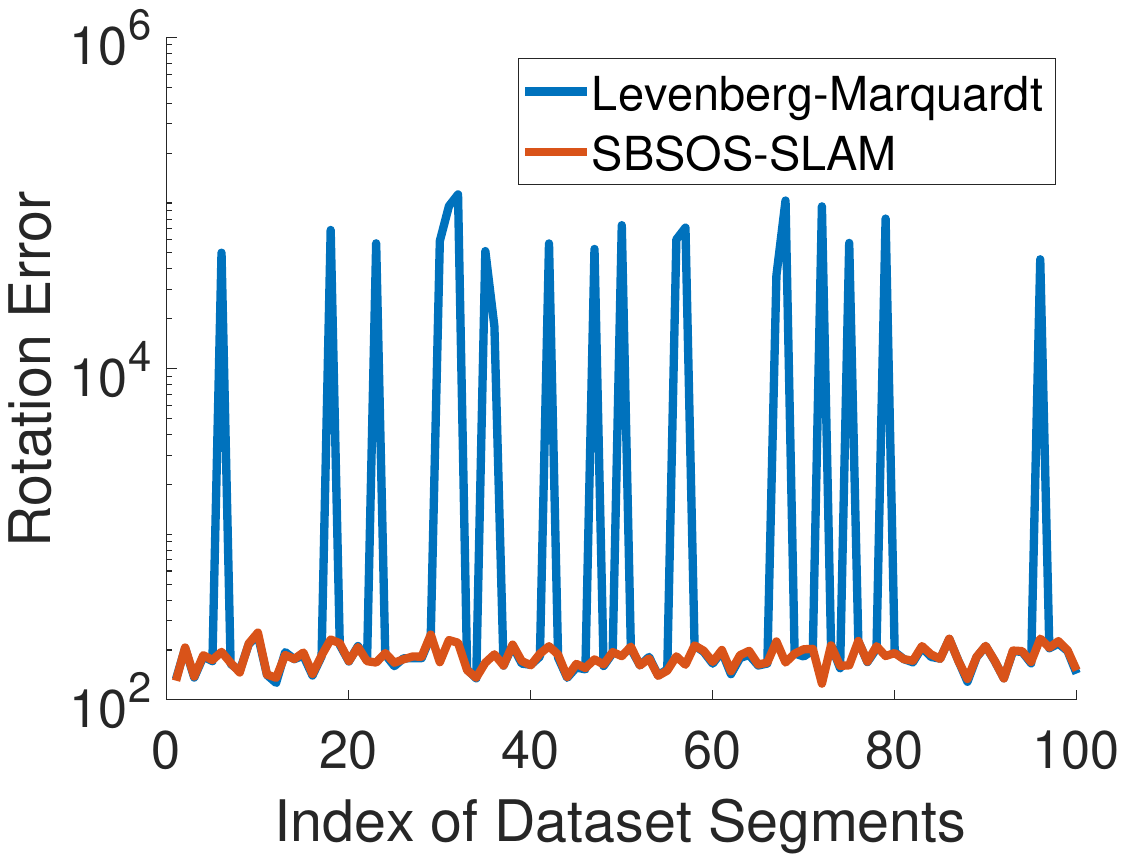}%
  \caption{\small Translational and rotational error verses groundtruth for the CityTrees10000 dataset.}%
  \label{fig:city_trees_error}%
\end{figure}

\begin{figure}[h!]%
  \centering%
  \includegraphics[trim={1cm, 6cm, 1cm, 6cm},clip,width=0.5\columnwidth]{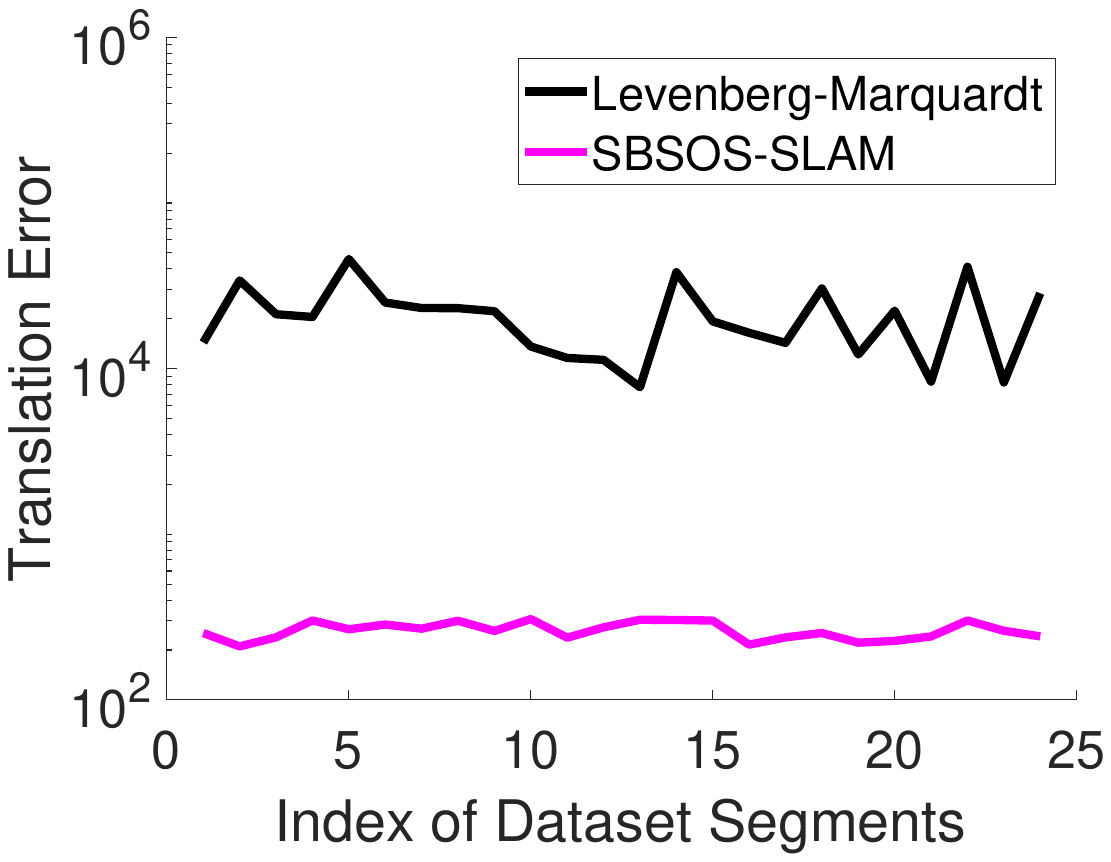}%
  \includegraphics[trim={1cm, 6cm, 1cm, 6cm},clip,width=0.5\columnwidth]{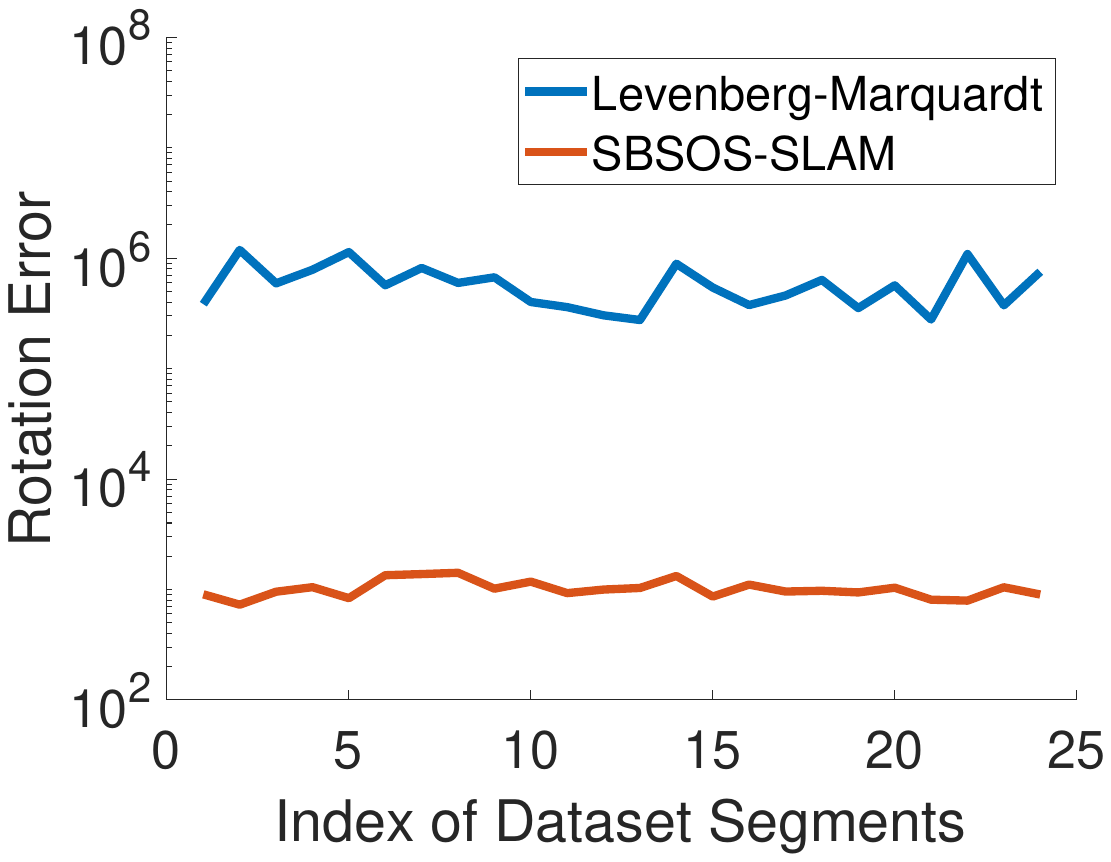}%
  \caption{\small Translational and rotational error verses groundtruth for the Manhattan3500 dataset.\vspace{-0.5cm}}%
  \label{fig:man_error}%
\end{figure}

We show several example plots where Levenberg-Marquardt gets stuck in a local minima, while SBSOS-SLAM is able to converge to ground truth at the first hierarchy and does not require initialization (\figref{fig:cityTrees100}, \figref{fig:man3500}).
\figref{fig:city_trees_error} and \figref{fig:man_error} show that our proposed algorithm
results in significantly smaller errors than Levenberg-Marquardt.

\subsection{SBSOS vs. Complex-domain SDP}

We compared the SBSOS and the Complex-domain SDP (CSDP) formulations of the Rotational Averaging Problem in terms of solving time and at what level of the hierarchy the problem converges numerically. 
Per \cite[Lemma 4]{weisser2018sparse} the SBSOS hierarchy convergences if the corresponding moment-based SDP, which is the dual of \eqref{eq:sbsos_hierarchy}, has a rank-1 solution. 
Convergence of the CSDP hierarchy is checked by comparing its result against the rank-1 solution of the SBSOS formulation. 

We randomly generated 500 fully connected graphs of 4,6,8,10 poses each.
We then applied the SBSOS and CSDP on these graphs and evaluated their performance at the first and second steps of the hierarchy.
Testing result are summarized in Table \ref{table: fully connected}.
The SBSOS and CSDP formulations converged at the second step of the hierarchy in all evaluated examples.
However, the solving time of CSDP at the second step of the hierarchy was considerably less than the solving time of SBSOS at the second step of its hierarchy.
In addition, more examples converged at the first step of the CSDP hierarchy when compared to the SBSOS hierarchy. 

\begin{table*}[]
\centering
\begin{tabular}{|c|c|c|c|c|}
\hline
\begin{tabular}[c]{@{}c@{}}Number \\ of poses\end{tabular} & \begin{tabular}[c]{@{}c@{}}SBSOS: 1st step\\ (Mean, Std, Opt)\end{tabular} & \begin{tabular}[c]{@{}c@{}}SBSOS: 2nd step\\ (Mean, Std, Opt)\end{tabular} & \begin{tabular}[c]{@{}c@{}}CSDP: 1st step\\ (Mean, Std, Opt)\end{tabular} & \begin{tabular}[c]{@{}c@{}}CSDP: 2nd step\\ (Mean, Std, Opt)\end{tabular} \\ \hline
4                                                      & (0.4638, 0.0577, 85.2\%)                                                   & (0.4887, 0.0471, 100\%)                                                    & (0.2840, 0.0367, 92.8\%)                                                   & (0.2836, 0.0367, 100\%)                                                    \\ \hline
6                                                      & (0.4003, 0.0212, 63.0\%)                                                   & (2.1716, 0.1657, 100\%)                                                    & (0.2600, 0.0157, 82.0\%)                                                   & (0.4220, 0.0200, 100\%)                                                    \\ \hline
8                                                      & (0.3909, 0.0216, 49.8\%)                                                   & (21.1445, 1.8422, 100\%)                                                   & (0.2434, 0.0155, 77.8\%)                                                   & (2.9081, 0.2239, 100\%)                                                    \\ \hline
10                                                     & (0.4310, 0.0406, 32.6\%)                                                   & (148.9397, 12.2855, 100\%)                                                 & (0.2954, 0.0251, 70.8\%)                                                    & (23.1850, 1.1476, 100\%)                                                   \\ \hline
\end{tabular}
\caption{Comparing the computation time and global convergence of the SBSOS and CSDP hierarchies on randomly generated fully connected pose graphs. 
Note that mean[sec] is the mean of solving time, std[sec] is the standard deviation of solving time, and opt is percentage of tests in which the problem converges.}
\label{table: fully connected}
\end{table*}

The two formulations were also tested using the Manhattan3500 dataset with 15, 20, 25, 30 and 50 poses 
In each test, the initial pose of the sub-graph of the dataset was randomly chosen.
The results for this test are summarized in Table \ref{table: man}.
Note that both SBSOS and CSDP formulations always converge by the second step in the hierarchy, and CSDP was considerably faster when compared to SBSOS.
Note that on these subgraphs problems that were not fully connected, both SBSOS and CSDP always converged at the first step of their respective hierarchies. 

\begin{table*}[]
\centering
\begin{tabular}{|c|c|c|c|c|c|}
\hline
\begin{tabular}[c]{@{}c@{}}Number \\ of poses\end{tabular} & \begin{tabular}[c]{@{}c@{}}Number \\ of tests\end{tabular} & \begin{tabular}[c]{@{}c@{}}SBSOS: 1st step\\ (mean, std, opt)\end{tabular} & \begin{tabular}[c]{@{}c@{}}SBSOS: 2nd step\\ (mean, std, opt)\end{tabular} & \begin{tabular}[c]{@{}c@{}}CSDP: 1st step\\ (mean, std, opt)\end{tabular} & \begin{tabular}[c]{@{}c@{}}CSDP: 2nd step\\ (mean, std, opt)\end{tabular} \\ \hline
15                                                     & 100                                                    & (0.2553, 0.0225, 100\%)                                                     & (0.9681, 2.2814, 100\%)                                                    & (0.1683, 0.0165, 100\%)                                                    & (0.2174, 0.1857, 100\%)                                                    \\ \hline
20                                                     & 100                                                    & (0.2225, 0.0262, 100\%)                                                     & (3.6315, 12.035, 100\%)                                                    & (0.1402, 0.0144, 100\%)                                                     & (0.4194, 0.9496, 100\%)                                                    \\ \hline
25                                                     & 100                                                    & (0.2323, 0.0335, 100\%)                                                     & (20.316, 83.996, 100\%)                                                    & (0.1482, 0.0248, 100\%)                                                     & (1.5447, 4.5637, 100\%)                                                    \\ \hline
30                                                     & 100                                                    & (0.2670, 0.0514, 100\%)                                                     & (27.278, 101.61, 100\%)                                                    & (0.2239, 0.0572, 100\%)                                                     & (3.2297, 12.423, 100\%)                                                    \\ \hline
50                                                     & 23                                                     & (0.7328, 0.4335, 100\%)                                                    & (527.13, 1006.4, 100\%)                                                    & (0.4806, 0.2420, 100\%)                                                    & (64.876, 140.94, 100\%)                                                    \\ \hline
\end{tabular}
\caption{Comparing the computation time and global convergence of the SBSOS and CSDP hierarchies on randomly selected subgraphs in the Manhattan3500 dataset.
Note that mean[sec] is the mean of solving time, std[sec] is the standard deviation of solving time, and opt is percentage of tests in which the problem converges.}
\label{table: man}
\end{table*}

 
\section{Conclusion}
\label{sec:conclusion}

In this paper, we proposed an algorithm called SBSOS-SLAM that formulates
the planar Pose Graph and Landmark SLAM problems as polynomial optimization
programs.
We also described how the same problem can be implemented in the complex domain as a hierarchy of semi-definite programs.
Empirical results showed that both formulations converged at the second step of the pair of hierarchies, and the implementation in complex domain was solved faster at higher hierarchical step compared to the SBSOS formulation. 



\balance
\bibliographystyle{style/IEEEtranN}
{\footnotesize
\bibliography{references/IEEEabrv,references/strings-short,%
    references/library}}

\begin{thebibliography}{32}
\providecommand{\natexlab}[1]{#1}
\providecommand{\url}[1]{#1}
\csname url@samestyle\endcsname
\providecommand{\newblock}{\relax}
\providecommand{\bibinfo}[2]{#2}
\providecommand{\BIBentrySTDinterwordspacing}{\spaceskip=0pt\relax}
\providecommand{\BIBentryALTinterwordstretchfactor}{4}
\providecommand{\BIBentryALTinterwordspacing}{\spaceskip=\fontdimen2\font plus
\BIBentryALTinterwordstretchfactor\fontdimen3\font minus
  \fontdimen4\font\relax}
\providecommand{\BIBforeignlanguage}[2]{{%
\expandafter\ifx\csname l@#1\endcsname\relax
\typeout{** WARNING: IEEEtranN.bst: No hyphenation pattern has been}%
\typeout{** loaded for the language `#1'. Using the pattern for}%
\typeout{** the default language instead.}%
\else
\language=\csname l@#1\endcsname
\fi
#2}}
\providecommand{\BIBdecl}{\relax}
\BIBdecl

\bibitem[Cadena et~al.(2016)Cadena, Carlone, Carrillo, Latif, Scaramuzza,
  Neira, Reid, and Leonard]{cadena2016past}
C.~Cadena, L.~Carlone, H.~Carrillo, Y.~Latif, D.~Scaramuzza, J.~Neira, I.~Reid,
  and J.~J. Leonard, ``Past, present, and future of simultaneous localization
  and mapping: Toward the robust-perception age,'' \emph{{IEEE} Trans. on
  Robotics}, vol.~32, no.~6, pp. 1309--1332, 2016.

\bibitem[Rosen et~al.(2016)Rosen, Carlone, Bandeira, and
  Leonard]{rosen2016sesync}
D.~Rosen, L.~Carlone, A.~Bandeira, and J.~Leonard, ``{SE-Sync}: A certifiably
  correct algorithm for synchronization over the {Special Euclidean} group,''
  in \emph{Proc. Int. Work. Algorithmic Foundations of Robot.}, 2016.

\bibitem[Carlone et~al.(2016)Carlone, Calafiore, Tommolillo, and
  Dellaert]{carlone2016planar}
L.~Carlone, G.~C. Calafiore, C.~Tommolillo, and F.~Dellaert, ``Planar pose
  graph optimization: Duality, optimal solutions, and verification.''
  \emph{{IEEE} Trans. on Robotics}, vol.~32, no.~3, pp. 545--565, 2016.

\bibitem[Rosen et~al.(2015)Rosen, DuHadway, and Leonard]{rosen2015convex}
D.~M. Rosen, C.~DuHadway, and J.~J. Leonard, ``A convex relaxation for
  approximate global optimization in simultaneous localization and mapping,''
  in \emph{Proc. {IEEE} Int. Conf. Robot. and Automation}, Seattle, Washington,
  USA, May 2015, pp. 5822--5829.

\bibitem[Carlone et~al.(2015)Carlone, Rosen, Calafiore, Leonard, and
  Dellaert]{carlone2015lagrangian}
L.~Carlone, D.~M. Rosen, G.~Calafiore, J.~J. Leonard, and F.~Dellaert,
  ``Lagrangian duality in {3D} {SLAM}: Verification techniques and optimal
  solutions,'' in \emph{Proc. {IEEE}/{RSJ} Int. Conf. Intell. Robots and
  Syst.}, Hamburg, Germany, September 2015, pp. 125--132.

\bibitem[Hu et~al.(2013)Hu, Khosoussi, and Huang]{hu2013towards}
G.~Hu, K.~Khosoussi, and S.~Huang, ``Towards a reliable slam back-end,'' in
  \emph{Proc. {IEEE}/{RSJ} Int. Conf. Intell. Robots and Syst.}, Tokyo, Japan,
  Nov 2013, pp. 37--43.

\bibitem[Liu et~al.(2012)Liu, Huang, Dissanayake, and Wang]{liu2012convex}
M.~Liu, S.~Huang, G.~Dissanayake, and H.~Wang, ``A convex optimization based
  approach for pose slam problems,'' in \emph{Proc. {IEEE}/{RSJ} Int. Conf.
  Intell. Robots and Syst.}, Vilmoura, Portugal, Oct 2012, pp. 1898--1903.

\bibitem[Kaess et~al.(2008)Kaess, Ranganathan, and Dellaert]{kaess2008isam}
M.~Kaess, A.~Ranganathan, and F.~Dellaert, ``{iSAM}: Incremental smoothing and
  mapping,'' \emph{{IEEE} Trans. on Robotics}, vol.~24, no.~6, pp. 1365--1378,
  2008.

\bibitem[Weisser et~al.(2018)Weisser, Lasserre, and Toh]{weisser2018sparse}
T.~Weisser, J.~B. Lasserre, and K.-C. Toh, ``{Sparse-BSOS}: a bounded degree
  {SOS} hierarchy for large scale polynomial optimization with sparsity,''
  \emph{Mathematical Programming Computation}, vol.~10, no.~1, pp. 1--32, 2018.

\bibitem[Brynte et~al.(2021)Brynte, Larsson, Iglesias, Olsson, and
  Kahl]{brynte2021tightness}
L.~Brynte, V.~Larsson, J.~P. Iglesias, C.~Olsson, and F.~Kahl, ``On the
  tightness of semidefinite relaxations for rotation estimation,'' \emph{arXiv
  preprint arXiv:2101.02099}, 2021.

\bibitem[Lasserre et~al.(2017)Lasserre, Toh, and Yang]{lasserre2017bounded}
J.~B. Lasserre, K.-C. Toh, and S.~Yang, ``A bounded degree {SOS} hierarchy for
  polynomial optimization,'' \emph{{EURO} Journal on Computational
  Optimization}, vol.~5, no. 1-2, pp. 87--117, 2017.

\bibitem[Durrant-Whyte and Bailey(2006)]{durrant2006a}
H.~Durrant-Whyte and T.~Bailey, ``Simultaneous localization and mapping
  ({SLAM}): {Part I},'' \emph{{IEEE} Robot. Autom. Mag.}, vol.~13, no.~2, pp.
  99--110, 2006.

\bibitem[Bailey and Durrant-Whyte(2006)]{bailey2006a}
T.~Bailey and H.~Durrant-Whyte, ``Simultaneous localization and mapping
  ({SLAM}): {Part II},'' \emph{{IEEE} Robot. Autom. Mag.}, vol.~13, no.~3, pp.
  108--117, 2006.

\bibitem[Thrun et~al.(2005)Thrun, Burgard, and Fox]{thrun2005probabilistic}
S.~Thrun, W.~Burgard, and D.~Fox, \emph{Probabilistic robotics}.\hskip 1em plus
  0.5em minus 0.4em\relax MIT press, 2005.

\bibitem[Lu and Milios(1997)]{lu1997a}
F.~Lu and E.~Milios, ``Globally consistent range scan alignment for environment
  mapping,'' \emph{Auton. Robot.}, vol.~4, no.~4, pp. 333--349, 1997.

\bibitem[Eustice et~al.(2005)Eustice, Walter, and Leonard]{eustice2005sparse}
R.~Eustice, M.~Walter, and J.~Leonard, ``Sparse extended information filters:
  Insights into sparsification,'' in \emph{Proc. {IEEE}/{RSJ} Int. Conf.
  Intell. Robots and Syst.}, Edmonton, AB, Canada, Aug 2005.

\bibitem[Dellaert and Kaess(2006)]{dellaert2006a}
F.~Dellaert and M.~Kaess, ``{Square Root SAM}: Simultaneous localization and
  mapping via square root information smoothing,'' \emph{Int. J. Robot. Res.},
  vol.~25, no.~12, pp. 1181--1203, 2006.

\bibitem[Boyd and Vandenberghe(2004)]{boyd2004convex}
S.~Boyd and L.~Vandenberghe, \emph{Convex Optimization}.\hskip 1em plus 0.5em
  minus 0.4em\relax New York, NY, USA: Cambridge University Press, 2004.

\bibitem[{Boumal}(2015)]{boumal2015arXiv_reimannian}
N.~{Boumal}, ``{A Riemannian low-rank method for optimization over semidefinite
  matrices with block-diagonal constraints},'' \emph{ArXiv e-prints}, Jun.
  2015.

\bibitem[Fan et~al.(2019)Fan, Wang, Rubenstein, and Murphey]{fan2019efficient}
T.~Fan, H.~Wang, M.~Rubenstein, and T.~Murphey, ``Efficient and guaranteed
  planar pose graph optimization using the complex number representation,'' in
  \emph{2019 IEEE/RSJ International Conference on Intelligent Robots and
  Systems (IROS)}.\hskip 1em plus 0.5em minus 0.4em\relax IEEE, 2019, pp.
  1904--1911.

\bibitem[Fan et~al.(2020)Fan, Wang, Rubenstein, and Murphey]{fan2020cpl}
------, ``Cpl-slam: Efficient and certifiably correct planar graph-based slam
  using the complex number representation,'' \emph{IEEE Transactions on
  Robotics}, vol.~36, no.~6, pp. 1719--1737, 2020.

\bibitem[Briales and Gonzalez-Jimenez(2017)]{briales2017cartan}
J.~Briales and J.~Gonzalez-Jimenez, ``{Cartan-Sync}: Fast and global
  {SE(d)}-synchronization,'' \emph{{IEEE} Robot. Autom. Letters}, vol.~2, 2017.

\bibitem[Lasserre(2009)]{lasserre2009moments}
J.~B. Lasserre, \emph{Moments, positive polynomials and their
  applications}.\hskip 1em plus 0.5em minus 0.4em\relax World Scientific, 2009,
  vol.~1.

\bibitem[Smail(2017)]{smail2017junction}
L.~Smail, ``Junction trees constructions in bayesian networks,'' in
  \emph{Journal of Physics: Conference Series}, vol. 893, no.~1.\hskip 1em plus
  0.5em minus 0.4em\relax IOP Publishing, 2017, p. 012056.

\bibitem[ApS(2015)]{mosek_c_api}
\BIBentryALTinterwordspacing
M.~ApS, \emph{The MOSEK C optimizer API manual Version 7.1 (Revision 54).},
  2015. [Online]. Available: \url{http://docs.mosek.com/7.0/capi/}
\BIBentrySTDinterwordspacing

\bibitem[Schmidt and Spitzer(1960)]{schmidt1960toeplitz}
P.~Schmidt and F.~Spitzer, ``The toeplitz matrices of an arbitrary laurent
  polynomial,'' \emph{Mathematica Scandinavica}, vol.~8, no.~1, pp. 15--38,
  1960.

\bibitem[Gregor(1988)]{gregor1988multidimensional}
J.~Gregor, ``The multidimensional $ z $-transform and its use in solution of
  partial difference equations,'' \emph{Kybernetika}, vol.~24, no.~7, pp. 1--3,
  1988.

\bibitem[Horn and Johnson(2012)]{horn2012matrix}
R.~A. Horn and C.~R. Johnson, \emph{Matrix analysis}.\hskip 1em plus 0.5em
  minus 0.4em\relax Cambridge university press, 2012.

\bibitem[ApS(2018)]{aps2018mosek}
M.~ApS, ``Mosek modeling cookbook,'' 2018.

\bibitem[Wolkowicz et~al.(2012)Wolkowicz, Saigal, and
  Vandenberghe]{wolkowicz2012handbook}
H.~Wolkowicz, R.~Saigal, and L.~Vandenberghe, \emph{Handbook of semidefinite
  programming: theory, algorithms, and applications}.\hskip 1em plus 0.5em
  minus 0.4em\relax Springer Science \& Business Media, 2012, vol.~27.

\bibitem[Serre(2002)]{Serre2002MatricesTA}
D.~Serre, ``Matrices: Theory and applications,'' 2002.

\bibitem[Olson et~al.(2006)Olson, Leonard, and Teller]{olson2006fast}
E.~Olson, J.~Leonard, and S.~Teller, ``Fast iterative alignment of pose graphs
  with poor initial estimates,'' in \emph{Proc. {IEEE} Int. Conf. Robot. and
  Automation}, Orlando, Florida, May 2006, pp. 2262--2269.

\end{thebibliography}

\end{document}